\documentclass[letterpaper,twocolumn,10pt]{article}
\usepackage{usenix-2020-09}

\usepackage{tikz}
\usepackage{amsmath}

\usepackage{filecontents}

\usepackage[utf8]{inputenc} 
\usepackage[T1]{fontenc}    
\usepackage{hyperref}       
\usepackage{url}            
\usepackage{booktabs}       
\usepackage{amsfonts}       
\usepackage{nicefrac}       
\usepackage{microtype}      
\usepackage{xcolor}         
\usepackage{graphicx}
\usepackage{subcaption}
\usepackage{ulem}
\usepackage{multirow}
\usepackage{makecell}
\usepackage{comment}

\newcommand{\sys}[0]{ByteCheckpoint}

\newcommand{\eg}{{e.g.}}

\newcommand{\cwu}[1]{{\textcolor{orange}{CWu: #1}}}

\title{\sys{}: A Unified Checkpointing System for Large Foundation Model Development}

\begin{document}
\pagestyle{empty}
\author{
    \rm{
        Borui Wan$^{\text{1}, *}$ \enskip
        Mingji Han$^{\text{2}, *}$ \enskip
        Yiyao Sheng$^{\text{2}}$ \enskip
        Yanghua Peng$^{\text{2}}$ \enskip
        Haibin Lin$^{\text{2}}$ \enskip
        Mofan Zhang$^{\text{2}}$ \enskip
    }
    \\
    \rm{
        Zhichao Lai$^{\text{2}}$ \enskip
        Menghan Yu$^{\text{2}}$ \enskip
        Junda Zhang$^{\text{2}}$ \enskip
        Zuquan Song$^{\text{2}}$ \enskip
        Xin Liu$^{\text{2}}$ \enskip
        Chuan Wu$^{\text{1}}$ \enskip
    }
    \\
    \\
    {$^{\text{1}}$\textit{The University of Hong Kong}\enskip $^{\text{2}}$\textit{ByteDance}\enskip}
}

\maketitle
{\let\thefootnote\relax\footnote{{$^*$Equal contribution.} Accepted by NSDI'25. Code: \href{https://github.com/ByteDance-Seed/ByteCheckpoint}{ByteCheckpoint}.}}

\begin{abstract}
Checkpointing to preserve training states is crucial during the development of Large Foundation Models (LFMs), for training resumption upon various failures or changes in GPU resources and parallelism configurations.
In addition, saved checkpoints are dispatched to evaluation tasks or transferred across different training stages (e.g., from pre-training to post-training). 
All these scenarios require resharding distributed checkpoints from one parallelism to another.
In production environments, different LFMs are trained with various frameworks and storage backends, depending on model sizes and training scales.
A high-performance checkpointing system 
is needed to enable efficient checkpoint management at scale throughout the lifecycle of LFM development. 

We introduce \sys{}, an industrial-grade checkpointing system for large-scale LFM training.
\sys{} features: a parallelism-agnostic checkpoint representation 
that enables efficient load-time checkpoint resharding;
a generic checkpoint saving/loading workflow to accommodate multiple training frameworks and support different storage backends; 
full-stack optimizations to ensure high I/O efficiency and scalability; 
a suite of monitoring tools to streamline large-scale performance analysis and bottleneck detection. 
Compared to existing open-source checkpointing systems~\cite{torch-dcp, megatron-dcp}, \sys{} significantly reduces runtime checkpoint stalls, achieving an average reduction of 54.20$\times$.
For saving and loading times, \sys{} achieves improvements of up to 9.96$\times$ and 8.80$\times$, respectively.
\end{abstract}

\section{Introduction}
\label{sec:intro}

Large Foundation Models (LFMs) in language~\cite{gpt4,llama3.1, gemini1.5, deepseek-v3}, vision~\cite{diffusion, dit}, audio~\cite{bark} and other modalities are revolutionizing today's AI landscape, spawning a variety of downstream applications such as conversational agents~\cite{chatgpt, character-ai}, coding assistants~\cite{github-copilot}, painting tools~\cite{midjourney}, and video~\cite{sora} generators.

Unlike traditional deep learning model training, LFM training is significantly more complex.
It has multiple training stages, including pre-training~\cite{gpt3, megascale} and post-training~\cite{wei2021sft,rlhf, hybridflow, ppo-framework}.
Furthermore, evaluation tasks are integrated into these stages to effectively assess model quality. Beyond the complexity, the development of LFMs is notably resource-intensive and time-consuming, largely due to immense model sizes and massive training datasets (e.g., DeepSeek-V3 comprises 671 billion total parameters, and is pre-trained on 14.8 trillion tokens)~\cite{llama3.1, deepseek-v3}.
The scale of LFM training can even be up to running on 12,288 GPUs~\cite{megascale}.


As a fundamental technique for preserving training states,
checkpointing captures snapshots of these states and stores them in persistent storage to facilitate training resuming.
Additionally, during LFM training, checkpoints are required for concurrent evaluation tasks that continually assess model quality.
Another use-case of checkpoints is dispatching those snapshots from pre-training to downstream post-training tasks, such as Supervised Fine-Tuning (SFT) or reinforcement learning.
The complex development pipeline, enormous scale, and prolonged duration of LFM training present significant challenges to designing a highly efficient checkpointing system.

\textit{First}, efficient and unified checkpoint management is necessary throughout the life cycle of real-world LFM development.
During training, checkpoint resharding is commonly required. 
This procedure transforms saved distributed checkpoints so they can be correctly loaded into a new parallelism configuration that differs from the one used for their creation.
In pre-training, variations in parallelism occur when problematic machines are removed, the available GPU quota fluctuates~\cite{qsync,lyra}, or training configurations (e.g., context length) and system optimization techniques~\cite{megascale} (e.g., kernel fusion, computation and communication overlapping
) are adjusted.
Across different stages and tasks, parallelism varies according to the scales of resources and datasets in use.
Highly efficient checkpoint resharding is needed to minimize extra overhead and maximize the end-to-end Effective Training Time Ratio (ETTR, calculated as the ratio between the productive training time and the wallclock time of a job)~\cite{meta-cluster}.
Moreover, on an industrial AI platform, LFM training jobs are initiated by various internal users or cloud customers who may choose different training frameworks (e.g., Megatron-LM~\cite{megatron}, PyTorch FSDP~\cite{fsdp}, DDP~\cite{ddp}, and others~\cite{vescale}) and select storage backends such as local disk, Hadoop Distributed File System (HDFS), or Network-Attached Storage (NAS) for storing checkpoints according to job characteristics and personal preferences.
Crafting customized checkpointing modules and workflows to accommodate these ad-hoc implementations complicates system development and substantially increases maintenance costs. It is vital to provide generic workflows for different training frameworks and storage backends.

\textit{Second},
Substantial I/O operations are involved in the checkpointing system for saving distributed checkpoints into persistent storage and loading them back for model training under various scenarios.
To minimize I/O blocking time and promptly save training states to persistent storage, it is crucial to expedite the I/O workloads.
In addition, ensuring the scalability of the checkpointing system is also essential since modern LFM training typically operates at large scales.

Existing checkpointing systems~\cite{checkfreq, check-n-run, gemini-sys, jit-check} assume consistent parallelism, 
and fail to address the demands for checkpoint resharding.
Although some efforts from the open-source community are devoted to developing checkpointing systems capable of resharding, they have several limitations. 
Some perform resharding in an inefficient offline manner~\cite{deepspeed-ucp}, while others only support specific training frameworks and storage backends~\cite{torch-dcp, megatron-dcp}.
Moreover, without customized optimizations and deployment experience, they suffer from sub-optimal I/O performance and lack the scalability to support large-scale LFM training in real-world production.

This paper presents the design, implementation, and deployment experience of \sys{}, a checkpointing system crafted for LFM development.
\sys{} incorporates a unified architecture (Sec.~\ref{sec:arch}) featuring an automatic \textit{resharding-on-loading} mechanism for distributed checkpoints (load-time checkpoint resharding) and supports multiple training frameworks and storage backends.
It integrates full-stack I/O performance and scalability optimization techniques.

$\triangleright$ \sys{}'s checkpoint 
representation is decoupled from the specific parallelism adopted during training 
(Sec.~\ref{sec:represent}), enabling efficient load-time checkpoint resharding.
For model and optimizer state representation, 
we separate each tensor shard's metadata from its numerical values and consolidate all the metadata into one global file.
The metadata of a tensor shard includes its basic runtime, position, and storage information. 
For the representation of dataloader states, we divide them into \textit{replicated} and \textit{sharded} states,
storing sharded states in individual files while the replicated ones are only saved by the training worker whose global rank is 0.

$\triangleright$ \sys{} offers a generic workflow for LFM training tasks employing different training frameworks and various storage backends (Sec.~\ref{sec:workflow}).
It tailors a planner for each framework to generate unified saving/loading plans.
These plans are then passed to a framework and storage backend agnostic engine to execute the I/O tasks.
\sys{}'s workflow utilizes this isolation in architecture, executing the same saving/loading steps for users with different training frameworks and storage backends.

$\triangleright$ \sys{} implements multiple optimizations (Sec.~\ref{sec:io_optimization}) to enhance I/O performance, including balanced and zero-redundancy plan generation, fully asynchronous execution pipelines, and efficient irregular tensor processing. 
We share our engineering experience in optimizing storage systems to support massive I/O requests, optimizing collective communications to guarantee stability, and designing monitoring and visualization tools for performance analysis and bottleneck detection
(Sec.~\ref{sec:industrial_scale}).
This full-stack approach scales \sys{} to support the training of a 405B LFM on 8,960 GPUs while still maintaining high efficiency.

\sys{} is deployed on our industrial AI platform with tens of thousands of GPUs for various LFM training tasks, including pre-training and post-training of language models, multi-modal understanding, and generation models (\eg, for video generation). 
\sys{} demonstrates significant advantages over existing open-source checkpointing systems, including PyTorch DCP~\cite{torch-dcp} and Megatron Distributed Checkpoint~\cite{megatron-dcp} (MCP).
Compared to the baselines, \sys{} achieves improvements ranging from 12.13$\times$ to 161.50$\times$ in terms of checkpoint stalls reduction, making the end-to-end checkpoint saving and loading procedures 6.05$\times$ and 3.88$\times$ faster on average, respectively.

\section{Background and Motivation}
\label{sec:background}
\subsection{LFM development}\label{sec:bg_lfm}

 \begin{figure}[!t]
  \centering
  \includegraphics[width=0.95\linewidth]{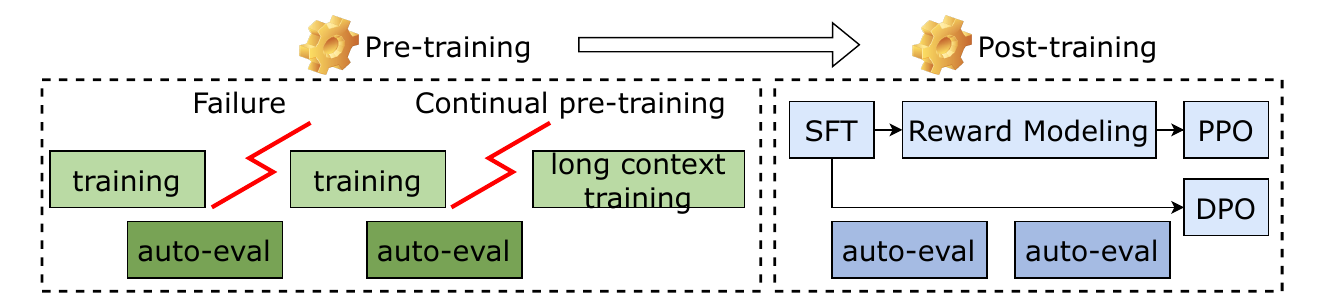}
  \caption{
  An overview of the training pipeline of LFM.
  }
  \label{fig:lfm_pipeline}
  \vspace{-3mm}
\end{figure}


\noindent \textbf{Production pipeline}.
As depicted in Fig.~\ref{fig:lfm_pipeline}, the development of LFM comprises pre-training and post-training stages.
During the initial pre-training phase, The LFM is iteratively trained on extensive data 
collected from multiple sources to absorb knowledge about the world. 
Subsequently, continual pre-training is employed to enhance the foundation model's capabilities. 
For instance, the pre-training of large language models (LLMs) typically involves long-context continual training to gradually increase the supported context length of LLMs. 
Post-training is employed to align the pre-trained model with human feedback or enhance the model's reasoning capabilities~\cite{openai-o1, deepseek-r1}.
Various task-specific labeled datasets (\eg, multilingual, code, math, reasoning, etc.) are involved in fine-tuning the LFMs, followed by reinforcement learning, typically including the reward modeling and then performing Proximal Policy Optimization~\cite{ppo} (PPO), or directly conducting Direct Preference Optimization~\cite{dpo} (DPO). 
Due to the reduced size of these datasets, relatively fewer GPUs are involved in post-training.
Automatic evaluation~\cite{characterization, check-n-run} is periodically triggered to get intermediate model checkpoints and assess the quality with diverse criteria.

\noindent \textbf{Checkpointing.}
Training states of LFM training jobs to be checkpointed include GPU and CPU states.
GPU states are learnable parameters in the LFM model and optimizer information (e.g., the float32 precision replica of the model and its momentum and variance in Adam~\cite{adam}). 
In 
state-of-the-art parallel training~\cite{megatron-lm3}, these states are sharded and placed across multiple GPUs. 
CPU states include 
 dataloader module, Random Number Generator (RNG) state, global training step number, and learning-rate scheduler, all stored in CPU memory. 
Our dataloader module incorporates a 
token buffer to cache input samples of varying lengths read from the data sources;
when the number of accumulated tokens reaches the context window size~\cite{megascale, llama3.1, opt, glm}, the dataloader assembles all cached samples into a batch (micro-batch). 
Due to the volatile nature of GPU and CPU memory, 
these training states should be periodically saved into persistent storage to tolerate any faults and prepare for evaluation tasks.

\noindent \textbf{Storage backends.} 
In production environments, 
separate distributed file systems (such as Tectonic~\cite{tectonic} for Llama 3.1 training~\cite{llama3.1}) are employed to store checkpoints~\cite{check-n-run, unicron} for formal tasks.
Given that various hardware failures and software bugs are inevitable during training~\cite{llama3.1, megascale,characterization}, storing checkpoints at different global training steps is necessary to safeguard training. 
Distributed file systems (e.g., HDFS, NAS) provide adequate storage capacity to accommodate multiple checkpoints of large models. 




\subsection{Checkpoint Resharding Scenarios}\label{sec:bg_ckpt}

\begin{figure}[!t]
  \centering
  \includegraphics[width=\linewidth]{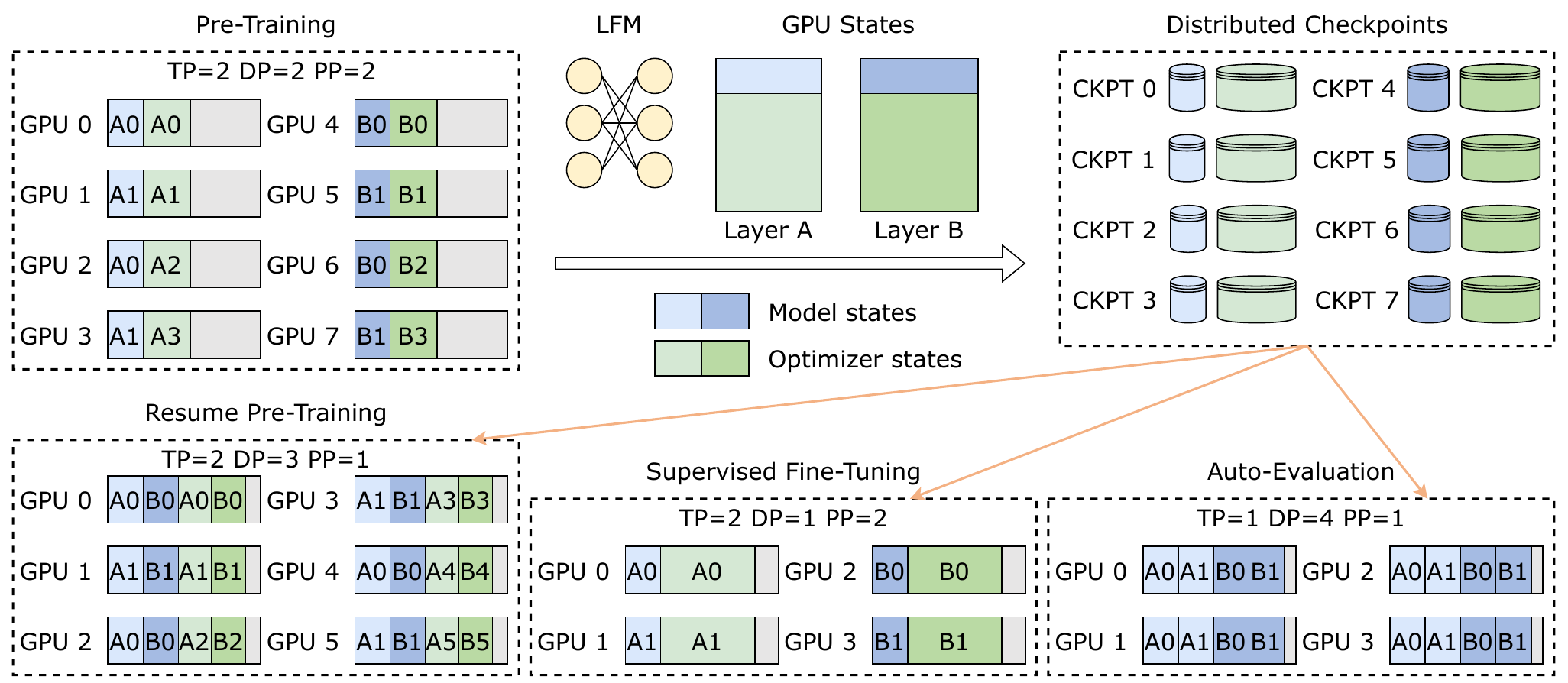}
  \caption{Checkpoint resharding scenarios in LFM training.
  We only show GPU states for clarity of the figure. 
  }
  \label{fig:reshard_req}
\end{figure}

During the life cycle of LFM development, checkpoint resharding is consistently required due to changes in parallelism under different scenarios (Fig.~\ref{fig:reshard_req}):

\noindent \textit{(1) Training Resumption}.
GPU quota allocated for an LFM training job can vary due to the removal of faulty machines~\cite{megascale, characterization}, adding new machines released from completed tasks, or running on tidal resources~\cite{parcae, qsync, lyra, antman}.
It is often necessary to adjust training parallelism in response to the resource changes, to maximize resource utilization.
Additionally, at the onset of large-scale pre-training, AI engineers often need to experiment with various model configurations and 
optimization techniques~\cite{megascale}, since conclusions from small-scale profiling or simulations do not always translate to optimal performance in large-scale training.
In this phase, adjusting parallelism configurations is common, resulting in frequent training resumption.
Moreover, long-context training changes the context length, which also requires GPU quota or parallelism adjustment.
As shown in the example in Fig.~\ref{fig:reshard_req}, 8 distributed checkpoint files are initially stored, then loaded into 6 training workers upon resuming.

\noindent \textit{(2) Cross-Stage Transition}.
When entering 
post-training 
, the number of GPUs involved typically decreases due to reduced training data in the latter.
Checkpoints saved from pre-training are frequently resharded to align with the specific workloads of each post-training task.
As depicted in Fig.~\ref{fig:reshard_req}, only 4 GPUs are involved for a fine-tuning task in the post-training stage, so the checkpoints are resharded accordingly.

\noindent \textit{(3) Evaluation}.
Evaluation tasks in both stages require loading model checkpoints and conducting inference on separate resources;
their parallelism needs to be adjusted to align with assigned GPUs used for specific datasets.
Fig.~\ref{fig:reshard_req} shows a 4-GPU eval task reshards model checkpoints from pre-training.

We collected the number of times of checkpoint resharding demands on our AI platform (training text, image, and video generation models) over the past six months and identified 1,870 instances of checkpoint resharding during pre-training resumption, 13,080 due to cross-stage reconfiguration, and 19,844 for running evaluation tasks.

\subsection{LFM Checkpointing Requirements}\label{sec:bg_challenges}
\noindent \textbf{Load-time checkpoint resharding}.
As a critical step in the life cycle of LFM development, the efficiency of checkpoint resharding is vital to minimize extra overheads.
In our AI platform, the previous common practice is to develop 
offline checkpoint resharding scripts and adapt the scripts whenever a new checkpoint resharding scenario arises.
This approach is inefficient and labor-intensive (see Appendix~\ref{sec:reshard_scripts} for further details).
In addition to the development costs, running resharding scripts leads to a substantial waste of GPU time and resources.
Table~\ref{tab:resharding_duration} presents the cost of executing resharding jobs in various scenarios.
Before training can resume or new evaluation tasks can begin, independent jobs that execute the resharding scripts must be submitted in advance.
These jobs download checkpoints from the storage systems, reshard distributed checkpoints to given parallelism configurations and upload new checkpoints back to the storage systems.
The targeted training or evaluation jobs cannot be executed until the completion of resharding jobs, resulting in prolonged pending time.
Moreover, since checkpoints created by resharding scripts are coupled with specific parallelism, they cannot be reused freely, which in turn increases the storage overhead.

Instead of executing scripts in independent jobs for checkpoint resharding, \textit{load-time checkpoint resharding}~\cite{torch-dcp} (also known as online  resharding) is preferred for LFM training.
It automatically identifies and retrieves necessary data from existing distributed checkpoints during the loading procedure.
To achieve this, the storage representation of checkpoints should be designed to be independent of specific parallelism. 
\begin{table}[!t]
\caption{
Average completion time of executing offline resharding jobs under different scenarios.
}
\label{tab:resharding_duration}
\resizebox{\linewidth}{!}{
\begin{tabular}{ccc}
    \toprule
    \textbf{Training Resumption} &  \textbf{Cross-Stage Transition} & \textbf{Evaluation}   \\ 
    \midrule
     1870.38s & 650.34s & 593.21s\\
    \bottomrule
\end{tabular}}
\end{table}

\begin{table}[!t]
\caption{Top three training frameworks used on our platform
}
\label{tab:framework_distribute}
\resizebox{\linewidth}{!}{
\begin{tabular}{cccc}
    \toprule
    \textbf{Framework} & \textbf{Pre-training} &  \textbf{Post-training} & \textbf{Average \#GPUs Per Job 
    } \\ 
    \midrule
    Megatron-LM & 13727& 68621& 301\\
    FSDP & 16842 & $\dagger$ &  25\\
    DDP & 25393& $\dagger$&  6\\
    \bottomrule
\end{tabular}}
\end{table}

\noindent \textbf{Multiple frameworks and storage backends}.
On our AI platform, a wide range of training frameworks are used, such as Megatron-LM~\cite{megatron}, DDP~\cite{ddp}, FSDP~\cite{fsdp}, veScale~\cite{megascale, vescale}, etc.
Table~\ref{tab:framework_distribute} lists the top three preferred training frameworks on our platform, based on six months of trace analysis. 
Users typically adopt Megatron-LM~\cite{megatron} for training large language foundation models.
FSDP~\cite{fsdp} is preferred for training tasks involving text-to-video or text-to-speech models, 
and DDP~\cite{ddp} is commonly used to train image encoder components of multimodal foundation models or for routine algorithm testing.
In addition, 
users can choose from various storage backends for checkpoint persistence depending on the scenario, ranging from local disks for debugging to HDFS or NAS for formal training tasks.
Each training framework comes with its own checkpoint module, file format, and ad-hoc implementation of the save/load logic.
However, these modules lack critical features for production, such as load-time resharding, asynchronous checkpointing, and support for remote persistent storage.
Integrating these features for each framework's checkpoint module and tailoring optimized implementations requires repeated engineering efforts.
This leads to inconsistent checkpointing interfaces across training frameworks and storage backends, complicating the codebase.
Moreover, the maintenance of diverse checkpoint file formats from different frameworks adds to the complexity of implementing checkpoint transfer logic across training stages and deploying models for evaluation and inference tasks.     
Therefore, it is vital to provide generic workflows for different training frameworks and storage backends. 


\begin{figure}[!t]
  \centering
  \includegraphics[width=\linewidth]{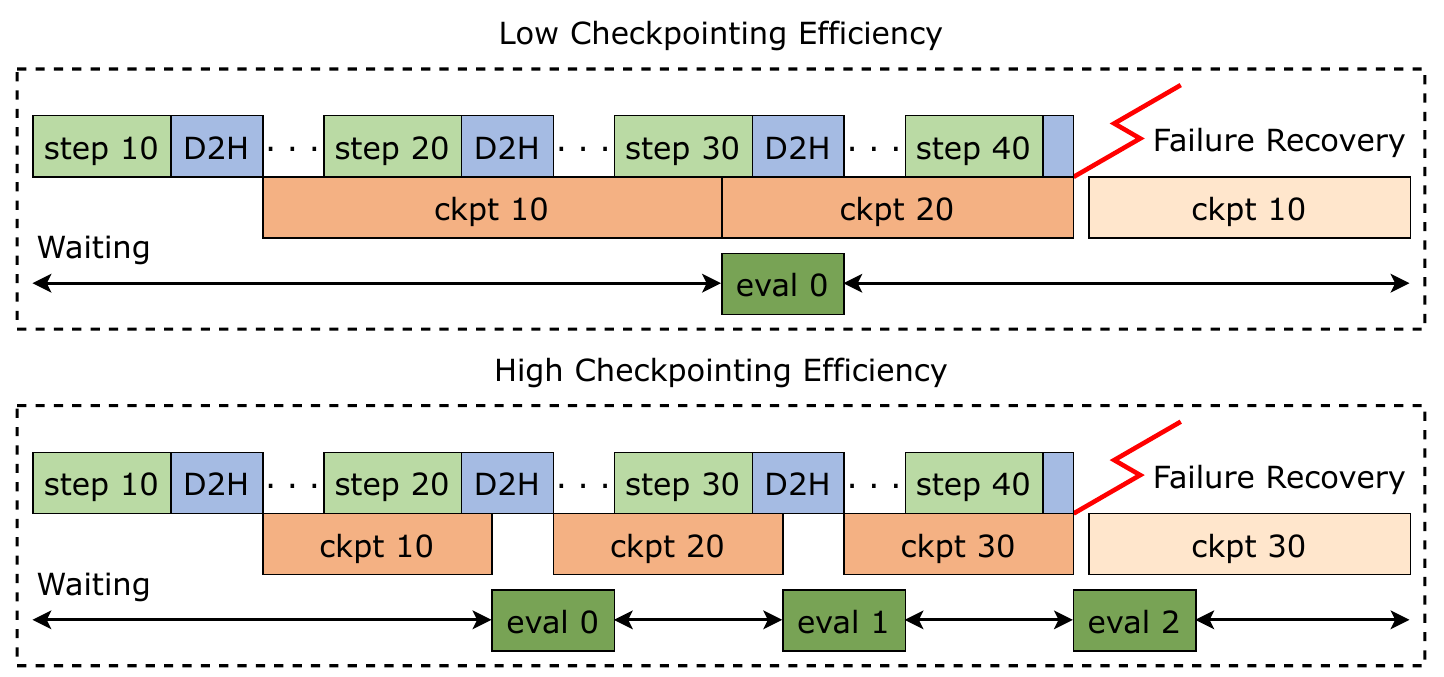}
  \caption{
  Checkpointing efficiency impacts failure recovery and evaluation tasks.
  D2H denotes the Device-to-Host copy.
  }
  \label{fig:check_efficiency}
  \vspace{-5mm}
\end{figure}

\noindent \textbf{Efficient and scalable I/O performance}.
Mainstream LFMs feature extensive model sizes, reaching hundreds of billions of parameters
~\cite{llama3.1}.
Consequently, the size of training states also increases significantly, imposing substantial overhead for checkpoint saving and loading.
By analyzing our previous LFM training jobs, we observe that 
the average end-to-end time required to save 
checkpoints of a GPT 175B model, trained on 4096 GPUs, to HDFS can be 200 seconds.
This duration substantially exceeds the time required for a single training iteration.
Even though this time-consuming procedure can be partially removed from the critical path of model training by adopting asynchronous checkpointing~\cite{checkfreq, megascale, reft, datastates-llm}, expediting end-to-end checkpointing remains crucial to minimize training progress loss~\cite{gemini-sys} caused by inevitable frequent failures in large-scale training~\cite{megascale}. 
As depicted in Fig.~\ref{fig:check_efficiency}, although checkpointing overlaps with training, its rapid completion allows more intermediate checkpoints to be stored before a failure occurs, enabling resumption from more recent states and improving ETTR.
Besides, evaluation tasks are triggered during training, and intermediate checkpoints are periodically pulled for these tasks.
Faster checkpointing ensures their timely execution, reducing blocking time due to preparing these checkpoints in remote persistent storage. 

Scaling the checkpointing system while maintaining the high I/O performance is non-trivial.
Sub-optimal and risky designs, which are hard to detect in small-scale settings, can lead to severe performance bottlenecks or even catastrophic job failures in large-scale training.
For example, massive read/write requests for checkpoint files from the training cluster to the storage systems (e.g., HDFS) can overload the master node, causing delays in file metadata operations.
Additionally, naive implementations of collective communications, such as integrity-checking barriers, introduce substantial initialization and synchronization overheads.
These overheads can even result in communication timeouts, ultimately causing the entire training job to fail.
Furthermore, as training scales up, efficiently analyzing system performance and detecting errors among training and I/O workers across multiple machines becomes increasingly challenging.

Existing checkpointing systems, such as CheckFreq~\cite{checkfreq}, Check-N-Run~\cite{check-n-run}, and Gemini~\cite{gemini-sys}, operate under the assumption of consistent parallelism and do not address the need for checkpoint resharding.
DCP~\cite{torch-dcp} and MCP~\cite{torch-dcp} incorporate checkpoint resharding capabilities but are limited in terms of supported parallelism strategies and training frameworks, I/O performance, and scalability.
A comprehensive discussion with related works is provided in Appendix~\ref{sec:related_work}.
In \sys{}, we design decoupled storage representation for efficient load-time checkpoint resharding, propose generic saving/loading workflows to support multiple frameworks and storage backends, integrate full-stack optimizations to enhance I/O performance and share our experience in scaling the checkpointing system to support real-world LFM training.
 
\section{System Design}
\label{sec:sys}

\begin{figure}[!t]
  \centering
  \includegraphics[width=\linewidth]{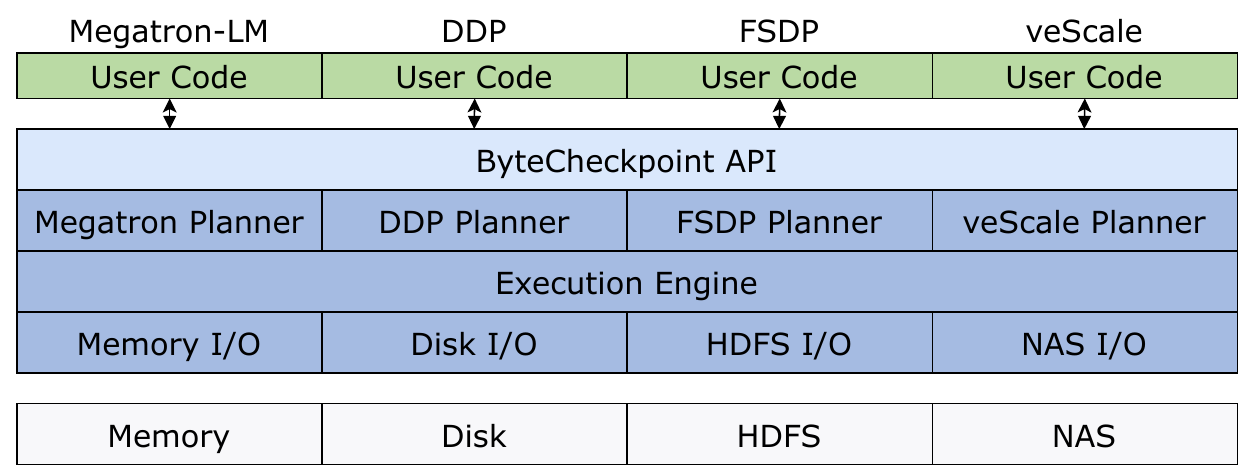}
  \caption{Architecture of \sys{}. 
  }
  \label{fig:architecture}
\end{figure}

Based on the above observations, we design \sys{} according to the following key principles:

\noindent \textit{(1) Decoupling}.
The checkpoint representation remains independent of specific runtime parallelism.
Interfaces of training frameworks and storage backends are separated from the core execution engine, ensuring robust extensibility.

\noindent \textit{(2) User-friendliness}.
The APIs should be concise, making it seamless for AI researchers and engineers to integrate them into their code and runtime environments.

\subsection{Unified Architecture}
\label{sec:arch}
The architecture of \sys{} is illustrated in Fig.~\ref{fig:architecture}.
Each component is introduced in detail as follows: 

\noindent \textbf{API.} \sys{}'s APIs (\texttt{bytecheckpoint.save} and \texttt{bytecheckpoint.load}) provide a unified entrypoint for user code across various training frameworks.
For instance, to save checkpoints, users first prepare the corresponding training states, checkpoint path, framework name, and performance-related options, then call \texttt{bytecheckpoint.save}.
This high-level entrypoint abstracts underlying system complexities, such as sharding specification, save/reshard plan generation, and I/O operations.



\noindent \textbf{Planner.}
The Planner serves as the interface for training frameworks.
It receives arguments (training states, checkpoint path, etc.) from the API layer, creates ShardMeta (Sec.~\ref{sec:represent}) for each tensor shard based on the worker's rank and framework-specific sharding specification such as \textit{Megatron ShardedTensor} or \textit{FSDP DTensor}, and determines the tensors and other states to save/load for each worker.
Each worker leverages the Planner to initially create local plans and subsequently collaborates to create global plans.
We implement a tailored planner for each training framework to extract information from these specifications and generate plans.


\noindent \textbf{Execution Engine}.
The Engine, running on each training worker, executes the saving/loading plans generated by the Planner when the respective API is called. It analyzes the given checkpoint path to determine the appropriate storage backend, then interacts with the Storage I/O layer to execute the actual I/O tasks specified in the plans.


\noindent \textbf{Storage I/O}.
Analogous to the design of the Planner layer, the Storage I/O layer encapsulates different storage backends and manages backend-specific read/write operations and optimizations. 
The interface between the Engine layer and the Storage I/O layer remains unified across different storage backends, facilitating the seamless integration of new storage backends.
\sys{} supports multiple storage options, including in-memory checkpoint storage~\cite{gemini-sys}, local disk storage, and remote storage systems.

\begin{figure}[!t]
  \centering
  \includegraphics[width=\linewidth]{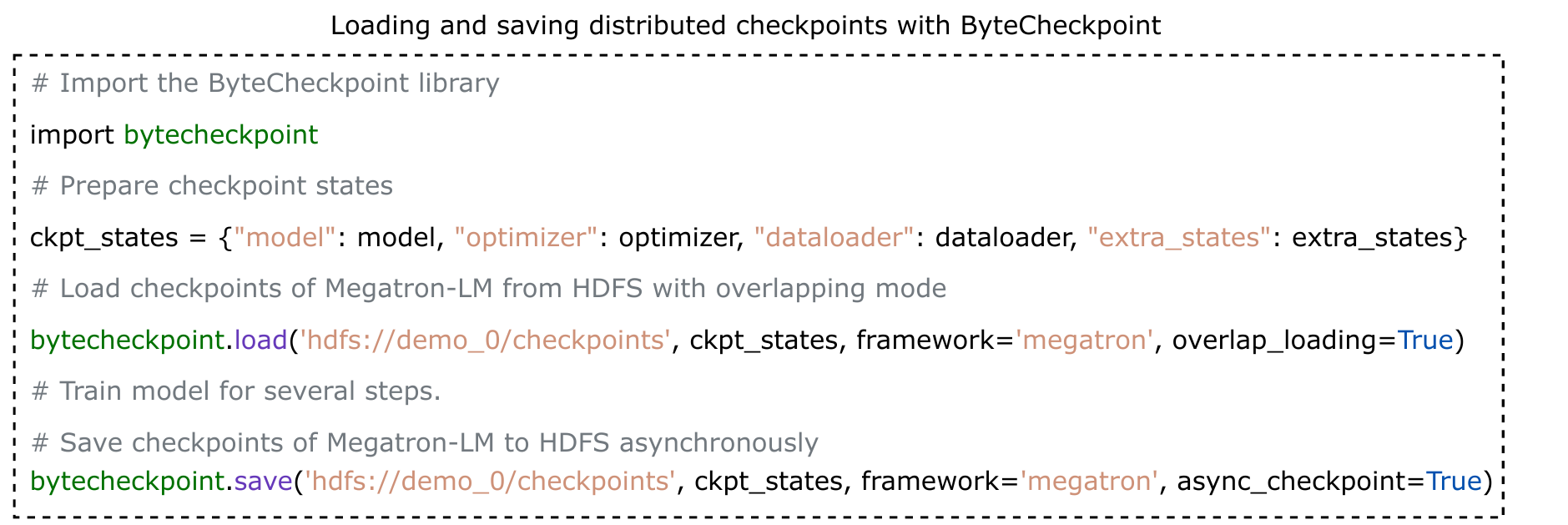}
  \caption{Examples of using \sys{}'s APIs.
  }
  \label{fig:code_examples}
\end{figure}

Fig.~\ref{fig:code_examples} illustrates a typical use case of \sys{}. Initially, users define a dictionary specifying the states to be saved or loaded, encompassing the model, optimizer, dataloader, and additional states.
At the outset of training resumption, stage transition, or evaluation, \texttt{bytecheckpoint.load} is invoked to load saved checkpoints.
Checkpoint resharding occurs automatically during loading when there are changes in parallelism.
During training, \texttt{bytecheckpoint.save} is periodically invoked to save checkpoints.

\subsection{Decoupled Checkpoint Representation}\label{sec:represent}

To achieve parallelism-agnostic checkpointing to enable automatic load-time resharding, we develop a formal specification of model and optimizer tensors in distributed training. Our design conforms to the Distributed Tensor and Checkpoint concepts in the \textit{PyTorch Distributed} library~\cite{ddp}.

\noindent \textbf{Model and optimizer states.}
Each tensor is uniquely identified by a "fully qualified name" (FQN) and has a global shape, representing its original shape before sharding.
Tensors can be either sharded or replicated across ranks (aka training workers).
For sharded tensors, the specific shard held by a worker is determined by three factors: the parallelism (e.g., tensor parallelism) applied to the tensor, the sharding dimension, 
and the group rank within the corresponding parallelism group. Based on a tensor's FQN, global shape, and sharding specification, \sys{} creates the tensor shard metadata (ShardMeta) for each rank, which is utilized in checkpoint storage to represent the tensor shard, independent of parallelism.
More precisely, ShardMeta for a tensor shard is an index tuple \textit{(fqn, nD\_offsets, nD\_lengths)}, where \textit{nD\_offsets} and \textit{nD\_lengths} indicate the local shard's offsets and lengths along the original multi-dimensional axes in its global shape.

\begin{figure}[!t]
  \centering
  \includegraphics[width=\linewidth]{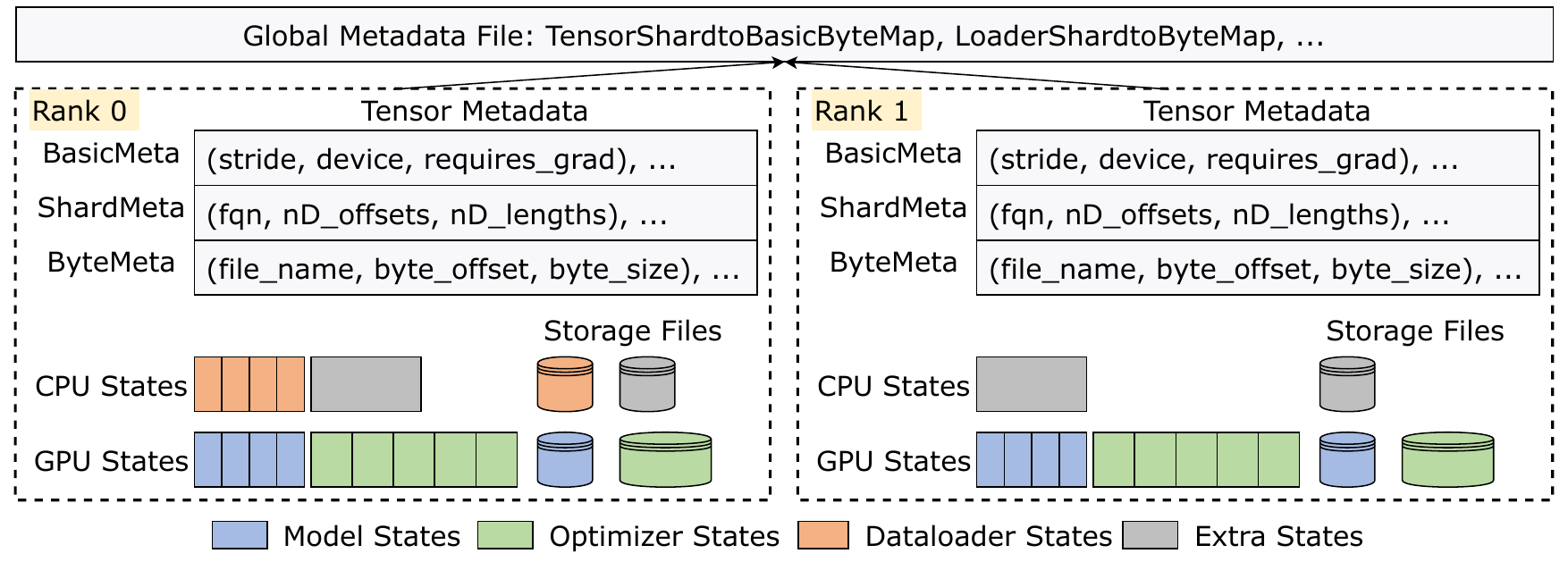}
  \caption{
  Checkpoint representation in \sys{}.
  The pipeline parallelism is applied in this example.
  }
  \vspace{-5mm}
  \label{fig:representation_meta}
\end{figure}

\noindent \textbf{Dataloader and extra states.}
The dataloader states can be categorized into two types: \textit{replicated} states and \textit{sharded} states. Replicated states include the number of data reading workers, paths to source datasets, and sampling ratios, and are identical across all I/O workers (subprocesses) in different ranks.
Sharded states are unique to each I/O worker, including the token buffer 
and data retrieval offsets for different data sources. 
In \sys{}, 
sharded states are saved in individual files, while replicated states are saved only by the I/O worker in rank 0. This 
reduces overall 
size of dataloader states to be saved and facilitates dataloader resharding when parallelism configurations change since the separation of data states into distinct files simplifies 
states merging and redistributing according to the new parallelism. For other extra states such as the RNG state, we pack and 
serialize them into one compact byte object before dumping them into storage.

\noindent \textbf{Checkpoint representation.}
Fig.~\ref{fig:representation_meta} illustrates the checkpoint representation in \sys{}.
Distributed checkpoints comprise a \textit{global metadata file}
and multiple \textit{storage files}.
Each rank 
generates three distinct files: a model state file, an optimizer state file, and an extra state file.
The Dataloader state file is generated only by training workers whose ranks for all parallelism degrees, except for DP degrees, are 0.
For tensor shards in model and optimizer states, their \textit{Metadata} consists of three parts:
\textit{BasicMeta}, which records essential information of individual tensor shards such as stride and device, critical for recovering the runtime state;
\textit{ShardMeta}, as previously introduced, recording relative position information of shards in the complete tensor;
\textit{ByteMeta}, which specifies the byte start offset and length of each tensor shard within the storage file.
All tensor metadata are consolidated into the global metadata file, and a mapping, termed \textit{TensorShardtoBasicByteMap}, is established between saved tensor shards and storage files based on the ShardMeta, BasicMeta, and ByteMeta of each tensor shard, ensuring accurate data retrieval.
The global metadata file also includes a \textit{LoaderShardtoByteMap}, which records the file index information of sharded states in each dataloader.
All storage files and the global metadata file are
stored in the specified storage backend designated by the checkpoint path.

\begin{figure}[!t]
  \centering
  \includegraphics[width=\linewidth]{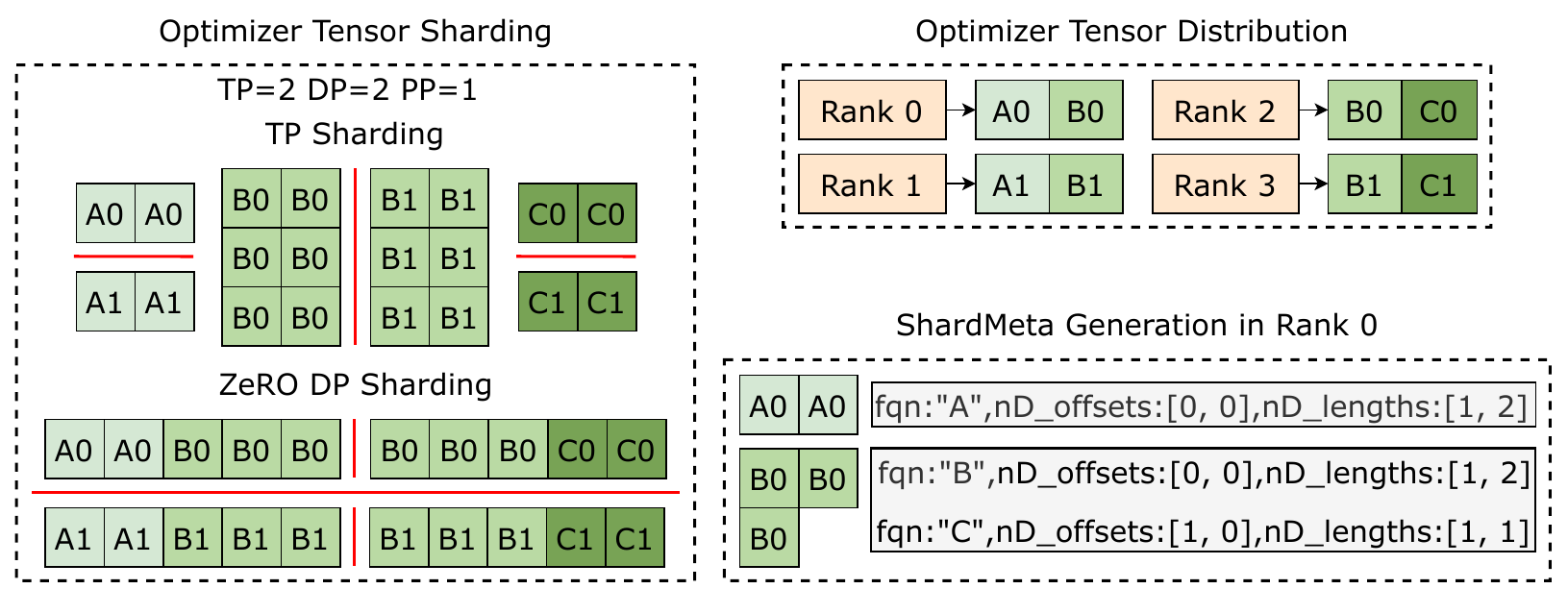}
  \caption{
  ShardMeta of irregular tensors in optimizer states.
  }
  \label{fig:irregular_decomposed_meta}
  \vspace{-5mm}
\end{figure}

\noindent \textbf{Decomposing irregular tensors}.
We define irregular tensors as those which, when flattened before sharding, cannot be reshaped back to their original dimensions. These tensors typically arise when applying the Zero Redundancy Optimizer (ZeRO)\cite{zero}, as implemented in Megatron-LM ZeRO2 and FSDP ZeRO3\cite{fsdp}. In these implementations, optimizer states within a DP group are flattened, concatenated, and sharded. Consequently, the resulting 1D tensor slices often cannot be directly represented using n-dimensional shapes and offsets.
Fig.~\ref{fig:irregular_decomposed_meta} illustrates an example of irregular tensor sharding: Unlike tensors $A$ and $C$, tensor $B$, with an original shape of (3, 2), is evenly split into two shards. Each shard, containing three elements, cannot be directly represented as a two-dimensional shape with corresponding offsets.
One intuitive approach to addressing the challenge of irregular tensor shards is to merge all tensor shards into complete tensors before saving checkpoints, thereby simplifying the generation of ShardMeta.
For example, to eliminate potential irregular tensors in DCP~\cite{torch-dcp}, FSDP performs synchronous all-gather communication operations, interleaved with D2H copy operations for each tensor shard, regardless of whether the shard is irregularly sharded.
However, this approach incurs significant communication overhead and requires frequent synchronization between GPU and CPU, substantially impeding efficiency.

To mitigate the overhead associated with merging tensor shards, \sys{} employs a tensor decomposition strategy for managing irregular tensor shards. Specifically, \sys{} decomposes an irregular tensor into a series of regular ones, representing each with index tuples. For example, consider the irregular tensor shard of tensor $B$ with three elements on rank $0$ in Fig.~\ref{fig:irregular_decomposed_meta}. \sys{} decomposes this shard into two regular tensors that can be directly represented by $nD\_offsets$ and $nD\_lengths$. This approach allows \sys{} to use multiple \textit{ShardMeta} entries to represent a single irregular tensor shard.
Although such decomposition slightly increases the metadata size and adds steps to the loading procedure, as reconstructing the target tensor may require looking up multiple smaller segments of irregular shards, it avoids the costly communication of tensor shards without extra blocking time during saving.

\subsection{Checkpoint Resharding 
Workflow}

\label{sec:workflow}

\begin{figure}[!t]
  \centering
  \includegraphics[width=\linewidth]{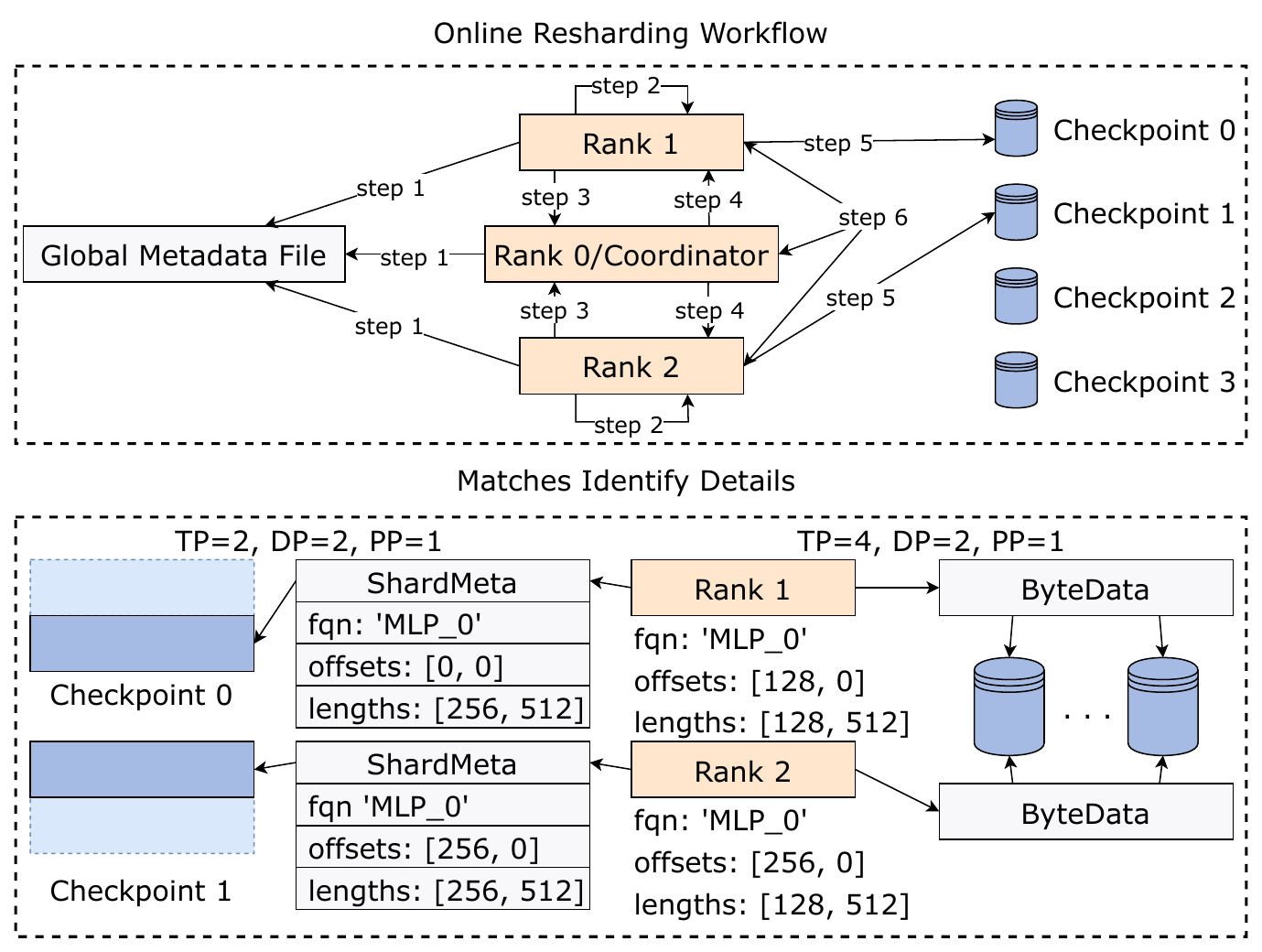}
  \caption{
  Load-time tensor resharding workflow.
  Checkpoint $k$ denotes the checkpoint saved by a previous live worker $k$.
  }
  \vspace{-5mm}
  \label{fig:tensor_reshard_example}
\end{figure}
%
\sys{} implements a generic workflow for saving and loading, automatically resharding checkpoints during the loading process. Leveraging the independence between the Engine Layer, Planner Layer, and Storage I/O Layer, we can execute consistent saving and loading steps across different training frameworks and storage backends. We take tensor resharding as an example
in Fig.~\ref{fig:tensor_reshard_example}, emphasizing how our checkpoint representation enables flexible load-time resharding.
The process for checkpoint saving and loading without resharding follows a similar procedure.

\noindent \textbf{Step 1}.
To initiate load-time checkpoint resharding, each rank invokes the \texttt{bytecheckpoint.load()} API, specifying the checkpoint path along with the model/optimizer
to be restored.
All ranks then load the global metadata file from the 
path.

\noindent \textbf{Step 2}.
For each tensor shard in the given model/optimizer, each rank queries the TensorShardToBasicByteMap within the global metadata file, identifying matching segments between the saved tensor shards and the sharding specification of new shards.
After identifying these matches, the planner constructs a local load plan, which includes BasicMeta and ByteMeta tailored to the specific shard. This identification mechanism is illustrated at the bottom of Fig.~\ref{fig:tensor_reshard_example}.

\noindent \textbf{Step 3}.
The coordinator planner, typically residing in rank 0, initiates a gather operation to aggregate loading plans from all ranks. It then optimizes each local plan by applying the redundant elimination optimization (Sec.~\ref{sec:opt_plan}) to distribute tensor shard loading workloads for reduced completion time.

\noindent \textbf{Step 4}.
The coordinator initiates a scatter operation to distribute the finalized loading plan. Each rank then receives its final loading plan from the coordinator.

\noindent \textbf{Step 5}.
The execution engine on each rank selects the storage backend wrapper in the Storage I/O layer based on the checkpoint path and then executes the loading pipeline (Sec.~\ref{sec:opt_pipeline}).

\noindent \textbf{Step 6}.
Upon completion of the loading pipeline, each rank utilizes the optimized asynchronous collective barrier primitive ensuring the atomicity of the distributed loading. Further details can be found in Appendix~\ref{sec:barrier_check}.

\begin{figure}[!t]
  \centering
  \includegraphics[width=\linewidth]{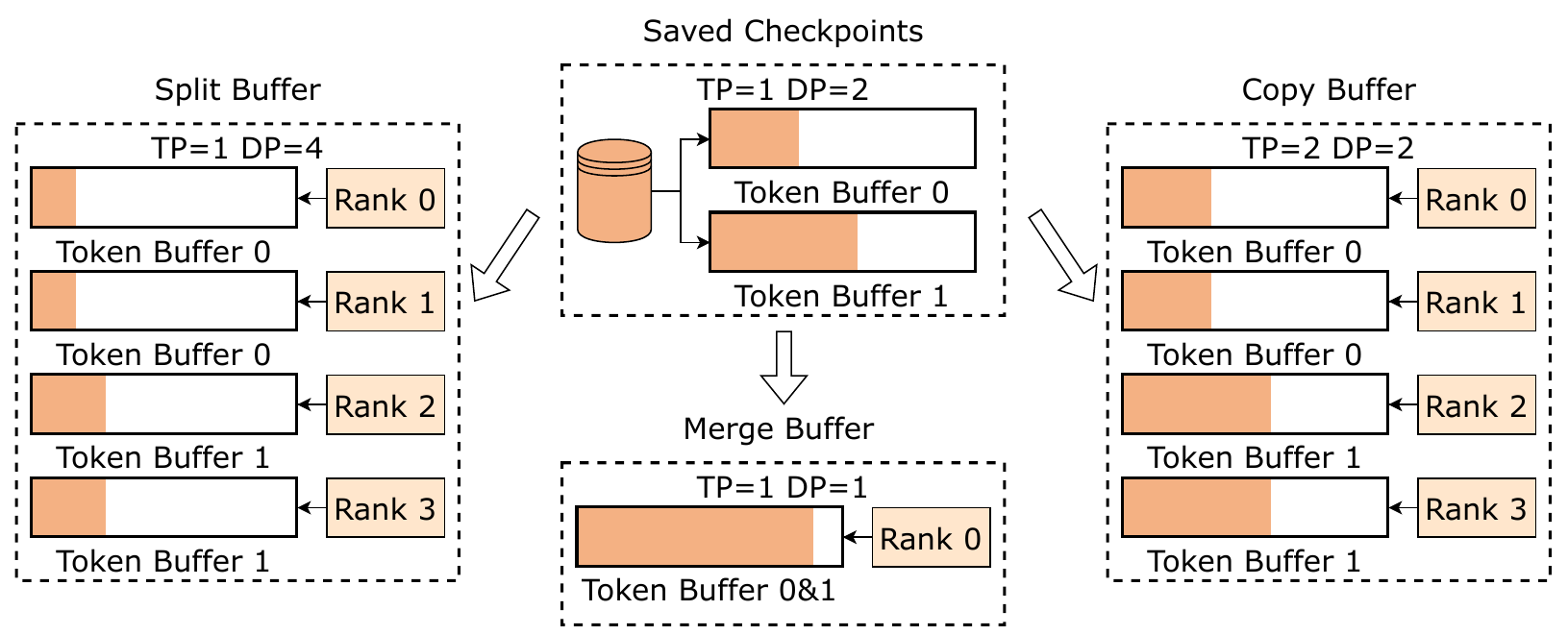}
  \caption{
  Examples of dataloader resharding.
  Sharded states such as cumulative token buffers are resharded according to the changes of 
  parallelism configurations.
  We do not depict data retrieval offsets for clarity of the figure.
  }
  \label{fig:loader_reshard_example}
\end{figure}

\noindent \textbf{Dataloader resharding}.
Similar to tensor resharding in Fig~\ref{fig:tensor_reshard_example}, \sys{} reshards the dataloader by querying the LoaderShardtoByteMap.
The illustration of dataloader resharding is depicted in Fig.~\ref{fig:loader_reshard_example}. When parallelism configuration changes, for the dataloader checkpoints, common items in the dataloader checkpoints can be loaded directly, whereas unique items such as accumulative token buffers and data retrieval offsets require resharding. Specifically, when the DP degree size remains constant while other parallel degrees are altered (TP degree changes in the example of Fig.~\ref{fig:loader_reshard_example}), the token buffers should be copied to the destination workers for bitwise-correct resuming. when there is a change in the DP degree size, the token buffers must be either split or merged accordingly to ensure that the resumed dataloaders do not discard cached data and do not retrain data that has already been sampled and fed. Thanks to the split storage strategy designed for the dataloader module, \sys{} can precisely identify the unique items that require resharding and process them efficiently.

\section{
Performance Optimization}
\label{sec:io_optimization}
We now elaborate on \sys{}'s performance optimization techniques, focusing on minimizing the overhead associated with checkpoint saving and loading.

\subsection{Optimized Plan Generation}
\label{sec:opt_plan}
\noindent \textbf{Balancing saving workload}.
In training scenarios employing data parallelism (DP), model states are replicated across all DP groups, resulting in duplicated model states. Existing checkpointing systems~\cite{torch-dcp, megatron-dcp} address this issue by designating the first DP group to save all model states. However, this approach leads to workload imbalance, potentially causing training workers in the first DP group to become stragglers.
To address this challenge, we implement a workload-balancing deduplication mechanism during the planning procedure, utilizing a Worst-Fit algorithm. Specifically, the coordinator planner distributes the saving workload based on the size of each tensor shard, assigning the current tensor shard to the rank with the smallest cumulative tensor shard size. This approach ensures a more equitable distribution of the saving workload across ranks.
 This balances workloads across all workers, improving saving efficiency.
 

\noindent \textbf{Eliminating redundant loading}.
When loading checkpoints into parallelism configurations that include data parallelism, \sys{} optimizes the process by eliminating repetitive tensor reading across DP groups, effectively combining storage file reading with tensor transferring. A crucial observation is that idle inter-GPU bandwidth can be utilized to concurrently transfer loaded tensors to peer GPUs during the file reading process. As illustrated in Fig.~\ref{fig:load_pipeline}, tensor reading workloads across DP groups are evenly distributed among training workers during the planning phase, thereby avoiding duplication of reads. Subsequently, I/O threads are launched to read the allocated tensor shards. Concurrently, in the main thread, shards read into CPU memory are first copied into GPU memory and then transferred to other workers that require them, utilizing all-to-all collective communication. 


\noindent \textbf{Plan and metadata cache}. In large-scale training, the execution of the planning procedure can introduce significant communication overhead, particularly when checkpoint saving occurs frequently.
For example, planning the saving procedure for a 405B transformer model distributed across 8960 GPUs requires 62 seconds.
However, we have observed that both the save plans and the global metadata file, although coupled with specific parallelism, remain constant throughout a single training session.
This allows for 
caching strategies to reduce planning times and associated overheads in subsequent checkpoint operations.
We introduce plan metadata caching, transforming the planning into a one-time cost.
Once established for the first time, the save plans and global metadata file are cached for future reuse,
eliminating repetitive planning.
\subsection{Fully Asynchronous Engine Pipeline}
\label{sec:opt_pipeline}




\begin{figure}[!t]
  \centering
  \includegraphics[width=\linewidth]{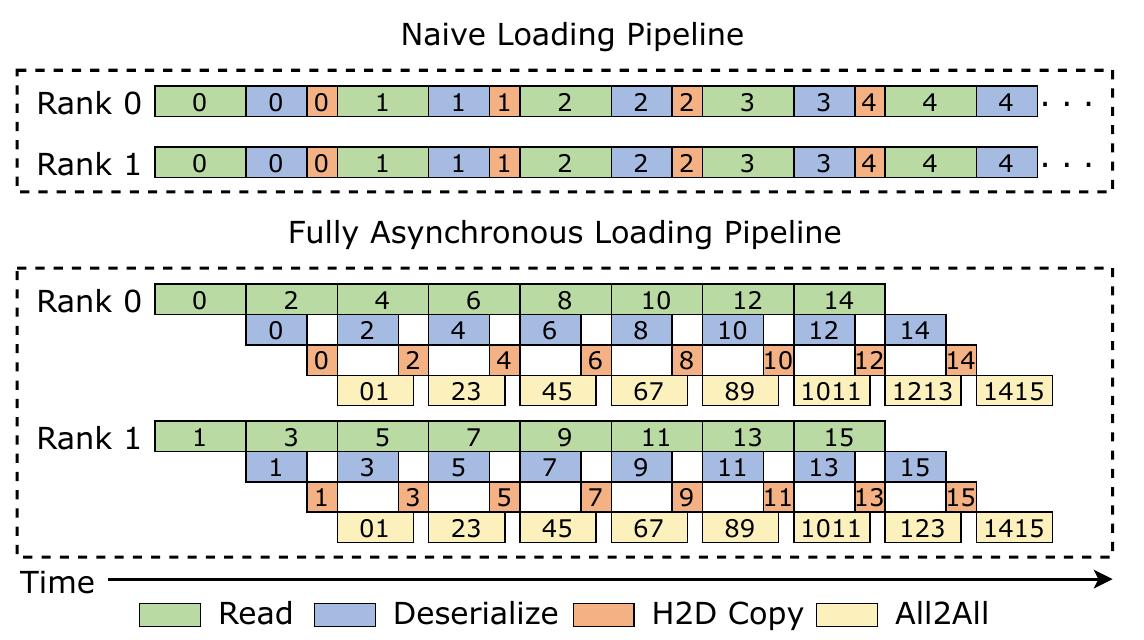}
  \caption{
   Comparison between \sys{}'s loading pipeline and the naive implementation.
  }
  \label{fig:load_pipeline}
\end{figure}


\sys{} Engine optimizes operation execution during checkpoint saving and loading (resharding) through pipelining. Using loading as an example (Fig.~\ref{fig:load_pipeline}), we pipeline file reading, deserialization, Host-to-Device (H2D) copy, and inter-GPU communication for each tensor shard, thereby achieving higher efficiency.
The \textit{read} operation downloads files containing desired tensors from the storage system and places them into shared memory (e.g., the \texttt{dev/shm} directory). The \textit{deserialize} operation deserializes tensors from shared memory. \sys{} employs multiple threads for parallel file downloading and deserialization. The \textit{H2D copy} operation transfers tensors from CPU memory to GPUs, while \textit{All2All} facilitates tensor transferring within the DP group.
For saving, we implement a symmetrical, fully asynchronous pipeline comprising 
D2H copy, serialization, and file uploading operations. To mitigate the performance impact of D2H copy on training, we employ a pinned CPU memory pool combined with a Ping-Pong buffering mechanism to accelerate this operation. We run multiple parallel processes to serialize tensors and dump files into shared memory.
Uploading threads actively monitor and initiate file uploading upon the completion of the dumping phase.

\subsection{High Performance Read/Write}\label{sec:io_parallel}
We optimize the read and write throughput of checkpointing by tailoring our implementation to optimal I/O 
usage of commonly used storage backends. For example, while HDFS is not primarily designed for random data access, it does offer some random read capabilities via its SDK, allowing applications to access specific file offsets and retrieve data from those positions. We leverage this feature 
and enable multi-threaded reading of a single file, significantly accelerating the download speed of checkpoint files from HDFS. In our production platform, single-file read speed has improved from 400 MB/s to 2-3 GB/s, all tested on H800 servers equipped with over 100 CPU cores, TB-level memory, and a 200Gbps NIC.

For file writing, HDFS's append-only write makes it impractical to split a single file into multiple parts based on offsets for multi-threaded writing. To overcome this limitation, we split the target file into several sub-files of fixed size and concurrently write them into HDFS using multi-threading.
After the upload is complete, we perform metadata-level concatenation to seamlessly merge the sub-files back into a single entity, ensuring the integrity of the stored data blocks. 
Without network congestion, single-file upload speed can reach 3 GB/s, far exceeding the average read/write throughput of a single HDFS client (under 100 MB/s~\cite{shvachko2010hdfs}). 

\subsection{Prefetching Dataloader States}
\label{sec:dataloader_prefetch}
Each dataloader employs multiple subprocesses, referred to as read workers,
to handle data loading and preprocessing.
When checkpointing is initiated, the main read worker
signals all 
other read workers to prepare their states. The training process is paused until all states are collected to ensure accuracy, as any update in the main process would result in the state change of workers.  
The duration of this blocking period depends on both the number of workers and the size of the accumulated token buffers.
For instance, when using a dataloader configured with 4 workers and a total state size of approximately 1GB, the state collection process typically takes around 8 seconds.

To alleviate this overhead, we adopt prefetching. 
Based on the pre-set checkpoint frequency, each read worker prepares its state during the training step just before checkpointing and puts the state into its state queue.
At the checkpointing step, the main read worker gathers those prepared states immediately through queue polling.
This optimization allows \sys{} to achieve near-zero dataloader state gathering delays.

\section{Checkpointing at Scale}
\label{sec:industrial_scale}
This section explores in depth the methods employed to optimize checkpointing for large-scale LFM training.

\subsection{Storage Support}\label{sec:scale_storage}

\noindent \textbf{High throughput, scalable storage system.} HDFS is the primary storage backend for \sys{}, we have implemented several optimizations to enhance throughput and scalability for large-scale checkpointing.  We have rewritten all HDFS components, including NameNode, DataNode, and SDK~\cite{hdfs}, in C++, effectively doubling performance compared to the original Java-based implementation.
Additionally, we have refined its architecture 
by introducing a new component named NNProxy~\cite{nnproxy}.
NNProxy functions as a stateless RPC proxy for HDFS NameNode, facilitating large-scale federated deployments of NameNodes while maintaining minimal querying latency.
The NNProxy design addresses the 
QPS bottleneck of HDFS NameNode metadata requests, a challenge intensified by the high volume of distributed checkpoints.
NNProxy also provides additional features such as authentication, rate limiting, and metadata query caching. These capabilities enable more granular management of checkpoints in production environments.
Our optimized HDFS achieves a substantial capacity of hundreds of PBs, 10 TB/s read/write bandwidth, and approximately 100,000 metadata QPS.

\noindent \textbf{Checkpoint cool-down strategy.} We have implemented a two-tier hot-cold storage architecture using a combination of SSD and HDD storage servers.
Our key observation is that newly stored checkpoint files are typically downloaded by evaluation tasks shortly after their creation.
However, 
where no training anomalies such as loss spikes occur,
the access frequency of these files decreases significantly after being downloaded by evaluation tasks.
Nonetheless, all checkpoint files must be preserved for traceability.
To efficiently manage storage space while ensuring data availability, we have developed a data cool-down mechanism that migrates data from SSD to HDD storage, thus ensuring that high-performance hot storage consistently has sufficient space for operations.
Specifically, we cool down all files that exceed the retention threshold based on their last modification time. We then remap the original file paths to the new HDD storage locations through pure metadata operations.
This strategy applies to HDFS directories and preserves the original access paths of the cooled-down files, providing a seamless user experience.


\subsection{Collective Communication}
\label{sec:opt_cc}

Collective communications (e.g., scatter, gather, barrier) are crucial in \sys{}'s workflow (Sec.\ref{sec:workflow}), particularly for planning and checkpoint integrity guarantee.
We initially used NCCL\cite{nccl} as the communication backend to execute gather and scatter operations at the coordinator. However, when scaling a pre-training task to 8960 GPUs, we observed that NCCL requires a long time to lazily build communication channels and allocate GPU memory during the planning stage of \sys{}'s saving workflow. In some cases, it becomes unresponsive or causes CUDA out-of-memory (OOM) errors, as scatter or gather operations necessitate establishing peer-to-peer communication with each GPU. These GPU
OOM issues and long initialization times are not apparent in small-scale trials but become significant at larger scales.

To enhance communication stability during planning, we have re-implemented the procedure using the gRPC framework~\cite{grpc}, which eliminates GPU memory usage during planning.
However, when scaling training to tens of thousands of GPUs, centralized gathering and scattering operations continued to impose a significant burden on the coordinator, leading to communication failures.
We further improved its stability by implementing a tree-based hierarchical communication topology.
Training workers on a single machine are organized into first-level subtrees, with the worker of local rank 0 designated as the root. For inter-machine communication, we iteratively group multiple machines, designating the worker with the lowest global rank in each group as the root. This procedure continues until all workers are integrated into a hierarchy converging at the global root (i.e., the coordinator).
In large-scale 3D parallel training scenarios, a TP-DP-PP communication tree naturally forms, removing extra connections. 

\subsection{Monitoring and Analysis}\label{sec:visualization}

\begin{figure}[!t]
  \centering
  \includegraphics[width=0.95\linewidth]{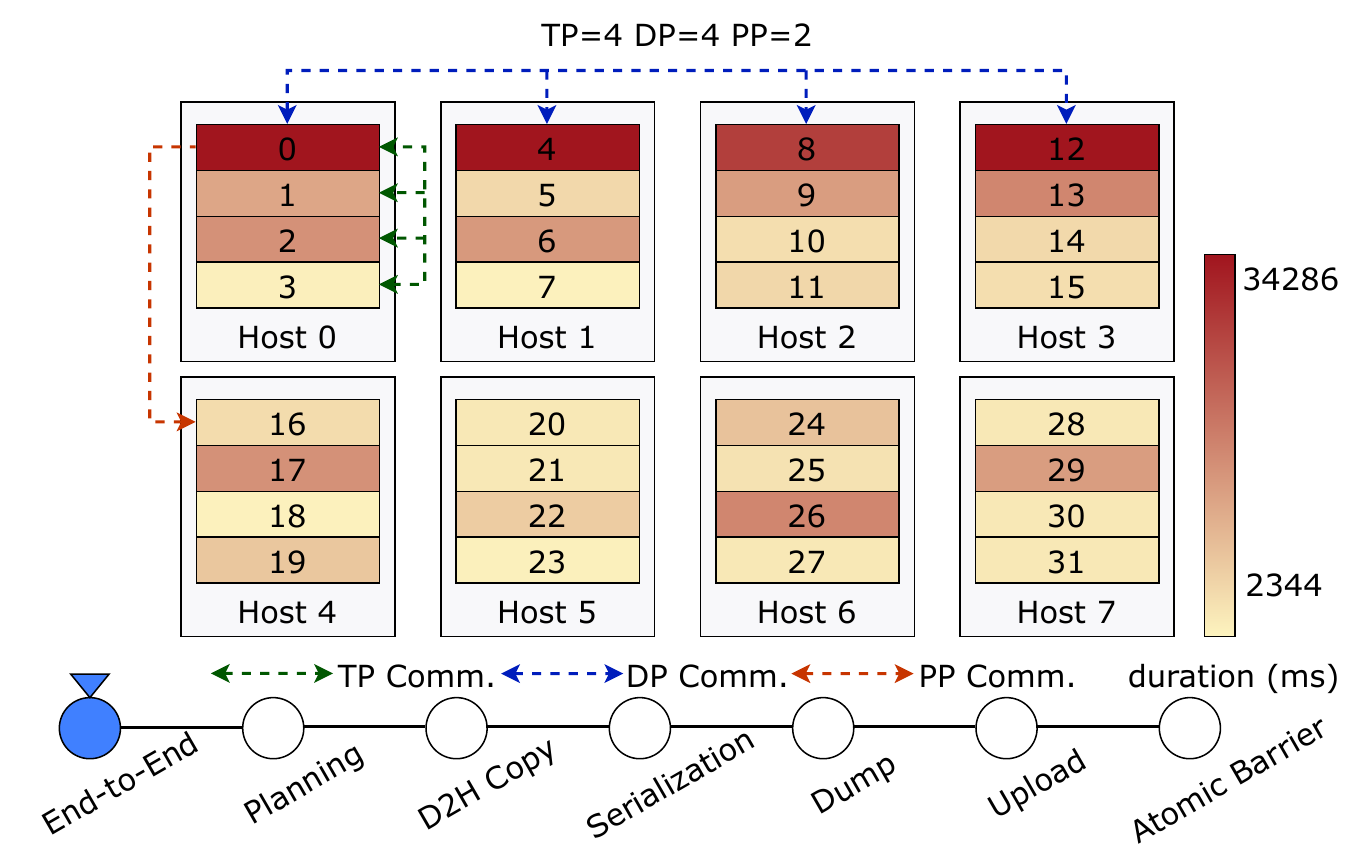}
  \caption{
  End-to-end checkpoint saving time heat map from a 3D parallel training task with Megatron-LM on 32 GPUs.
  The color denotes the time of a selected 
  phase on a rank. 
  }
  \vspace{-4mm}
  \label{fig:heatmap}
\end{figure}

\begin{figure}[!t]
  \centering
  \includegraphics[width=0.95\linewidth]{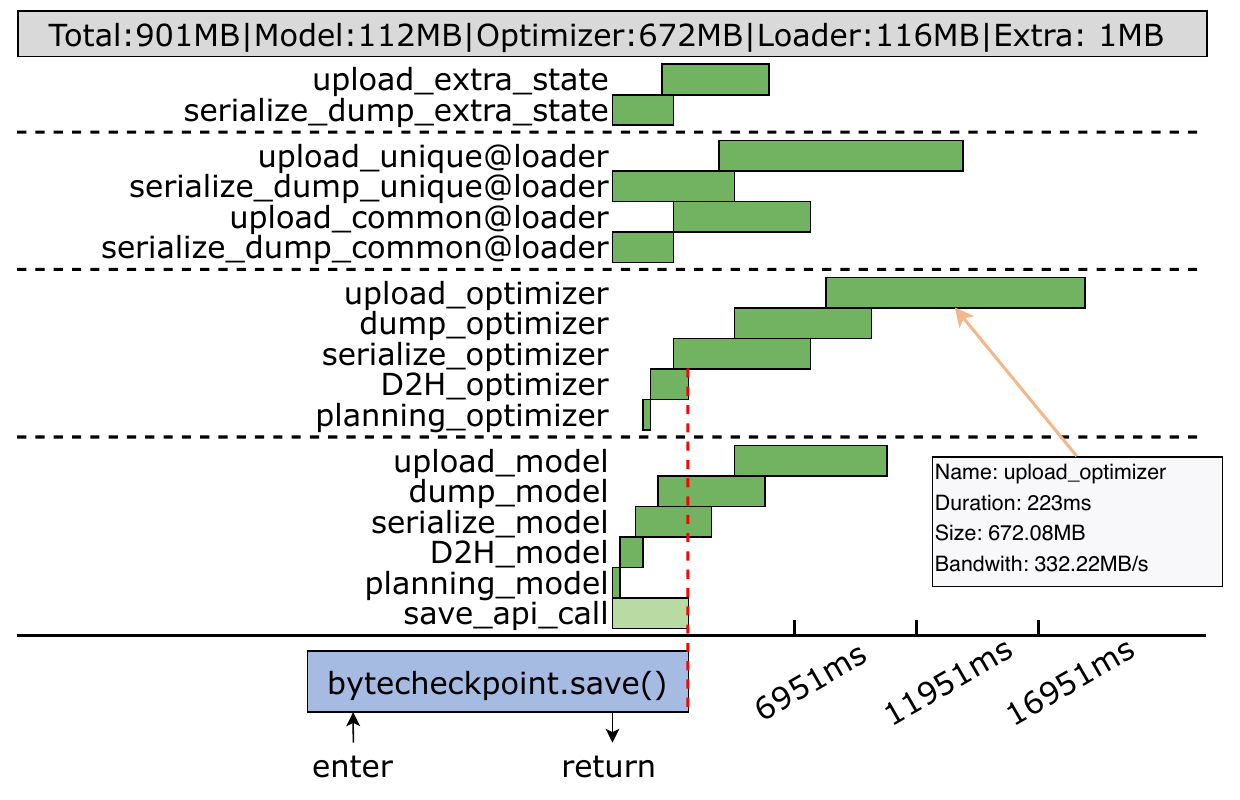}
  \caption{
    Time breakdown of checkpoint saving on rank 0.
  }
  \vspace{-5mm}
  \label{fig:breakdown}
\end{figure}

During training, \sys{} continuously collects critical performance measurements and visualize them for real-time performance monitoring and analysis. This approach enables quick checkpointing issues detection, such as low read/write throughput, stragglers, and upload failures and retries.

\noindent \textbf{Data collection}.
We have designed a user-friendly metrics system based on Python's context manager and decorator syntax to monitor critical procedures flexibly.
It automatically captures the duration and I/O size of each operation, along with relevant metadata such as each worker's rank, the file path, and the current step.
All collected metrics are transferred to a remote database via a background message queue.

\noindent \textbf{Visualization}. 
\sys{} provides users with a comprehensive topological performance overview of all ranks associated with different checkpoint phases, as well as detailed duration breakdowns for any specific rank.
Fig.~\ref{fig:heatmap} presents an exemplary heat-map visualization of checkpoint saving times within a 3D parallel training topology.
The visualization highlights that ranks 0, 4, 8, and 12 experience the longest saving times as their checkpoints including dataloader states. Additional metrics visualizations are also available for detailed analysis,
such as blocking time and fine-grained phases like planning and D2H copy. These visualizations enable users to easily pinpoint straggler nodes or phases for both system development and in production environments. For instance, if certain nodes experience network issues, increased upload or download times during the HDFS transfer phase would be readily apparent.
Furthermore, detailed timeline breakdowns of checkpointing procedures at each rank are accessible via the heat map topology overview, enabling users to thoroughly evaluate all optimizations in the system.
Fig. ~\ref{fig:breakdown} illustrates the execution details of each checkpoint-saving phase for a specific rank.


\noindent \textbf{Storage-side monitoring}.
On the storage client side, we monitor the latency and I/O size of each atomic read/write operation at the I/O chunk level.
All collected metrics are periodically transmitted via a message queue to a ClickHouse service, where the data are aggregated and analyzed.
Unexpectedly high latency or low bandwidth triggers alerts to engineers for further investigation.
On the storage cluster side, our primary focus is on overall performance metrics, including metadata request QPS, cluster-level read/write throughput, and storage capacity utilization.
This comprehensive monitoring system plays a crucial role in identifying issues such as slow checkpoint reads and writes, and even complete system unavailability due to storage capacity exhaustion, facilitating the implementation of preventive measures for enhancing system reliability and efficiency.



\begin{table*}[!t]
\caption{Model and parallelism configurations.
}
\label{tab:model_parallel_config}
\resizebox{\linewidth}{!}{
\begin{tabular}{ccccccccc}
    \toprule
    \textbf{Model} & \textbf{Hidden Size} & \textbf{\#Heads} & \textbf{\#Layers} & \textbf{\#Parameters} & \textbf{Source \#GPUs} & \textbf{Soure Parallelism} & \textbf{Target \#GPUs} & \textbf{Target Parallelism}\\ 
    \midrule
     \multirow{2}{*}{vDiT} &  \multirow{2}{*}{1664} &  \multirow{2}{*}{16} & \multirow{2}{*}{48} & \multirow{2}{*}{4B} & 32 & ZeRO-2 & 64 & ZeRO-2 \\
      & & & & & 128 & ZeRO-2 & 64 & ZeRO-2 \\
      \midrule
     \multirow{2}{*}{tGPT} &  \multirow{2}{*}{8192} & \multirow{2}{*}{64} & \multirow{2}{*}{80} &  \multirow{2}{*}{70B} & 2400 & TP=4, DP=75, PP=8 & 4800 & TP=4, DP=150, PP=8 \\
     & & & & & 4800 & TP=4, DP=150, PP=8 & 2400 & TP=4, DP=75, PP=8 \\
    \bottomrule
\end{tabular}}
\end{table*}

\begin{table*}[!t]
\caption{I/O performance comparison among \sys{}, DCP and MCP.}
\label{tab:io_compare}
\begin{small}
\resizebox{\linewidth}{!}{
\begin{tabular}{c|cc|ccccccc}
    \toprule
    \textbf{Model and Framework} & \textbf{Source \#GPUs} & \textbf{Method} & $\mathbf{T_{Block}}$ \textbf{(s)} & $\mathbf{T_{Save}}$ \textbf{(s)} & $\mathbf{T_{Load}}$ \textbf{(s)} & $\mathbf{T_{Reshard}}$ \textbf{(s)} & \textbf{ETTR (\%)}\\
    \midrule
     \multirow{2}{*}{vDiT 4B} & \multirow{2}{*}{32} & DCP & 16.25 & 86.82 & 50.12 & 74.89 & 38.60\\
     & & \sys{} (GPU states) & 0.54 (\textbf{30.09$\times$}) & 27.47 (\textbf{3.16$\times$}) & 11.69 (\textbf{4.29$\times$}) & 16.01 (\textbf{4.68$\times$}) & 46.22 (\textbf{1.20$\times$})\\
     \cline{2-8}
     \multirow{2}{*}{FSDP} & \multirow{2}{*}{128} & DCP & 61.37 & 236.34 & 105.74 & 91.01 & 41.62\\
     & & \sys{} (GPU states) & 0.38 (\textbf{161.50$\times$}) & 23.74 (\textbf{9.96$\times$}) & 12.01 (\textbf{8.80$\times$}) & 13.64 (\textbf{6.67$\times$}) & 48.92 (\textbf{1.18$\times$})\\
     \midrule
     \multirow{3}{*}{tGPT 70B} & \multirow{3}{*}{2400} & MCP & 4.73 & 28.97 & 69.87 & 126.30 & 34.61\\
     & & \sys{} (GPU states) & 0.39 (\textbf{12.13$\times$}) & 13.11 (\textbf{2.21$\times$}) & 49.48 (\textbf{1.41$\times$}) & 62.10 (\textbf{2.03$\times$}) & 40.18 (\textbf{1.16$\times$})\\
     & & \sys{} (full states) & 0.50 & 20.55 & 72.35 & 401.21 & 28.80\\
     \cline{2-8}
     \multirow{3}{*}{Megatron-LM} & \multirow{3}{*}{4800}  & MCP & 4.70 & 76.21 & 123.80 & 64.62 & 31.28\\
     &  & \sys{} (GPU states) & 0.36 (\textbf{13.06$\times$}) & 8.59 (\textbf{8.87$\times$}) & 64.39 (\textbf{2.00$\times$}) & 54.31 (\textbf{1.19$\times$}) & 40.29 (\textbf{1.29$\times$})\\
     &  & \sys{} (full states) & 0.43 & 25.04 & 83.56 & 115.95 & 34.70\\
    \bottomrule
\end{tabular}}
\end{small}
\end{table*}

\section{Evaluation}



In this section, we detail our deployment and operational experiences with \sys{}.
\sys{} is built upon DCP~\cite{torch-dcp} (commit hash: 80c07df), a state-of-the-art open-source checkpointing system that supports checkpoint resharding for FSDP~\cite{fsdp}. The implementation of 
\sys{} consists of about 20,000 lines of Python code.

\subsection{
Checkpointing Efficiency}
\label{sec:exp_main}


We demonstrate checkpointing 
performance of \sys{} relative to baselines~\cite{torch-dcp, megatron-dcp} with different workloads.

\noindent \textbf{Workloads}.
We adopt two different transformer-based~\cite{attention} structures, i.e. DiT~\cite{dit} and GPT-3~\cite{gpt3} to implement two models referred to as \textit{vDiT} and \textit{tGPT}, respectively.
The vDiT model is fine-tuned with the FSDP framework for video generation on a cluster of NVIDIA A100 80GB GPUs, while the tGPT model is trained for text generation with Megatron-LM~\cite{megatron} on NVIDIA H800 80GB GPUs.
Detailed configurations for model and parallelism are provided in Table~\ref{tab:model_parallel_config}, where the source number of GPUs and parallelism pertain to the saving and loading evaluations, while the target number of GPUs and parallelism correspond to the configurations used for load-time resharding evaluations. All machines in the training clusters are interconnected via InfiniBand,
with HDFS selected as the persistent storage solution for all experiments.
We integrate the test models into our LFM training trials, conducting training over 500 steps and saving checkpoints every 100 steps. We then resume training under two scenarios: with and without changes in GPU numbers and parallelism configurations, to assess the performance of load-time resharding and standard checkpoint loading.
Saving performance is evaluated based on average additional overheads, specifically checkpoint stalls (training blocking time) and end-to-end checkpoint saving time (from the save API call to the completion of checkpoint integrity checking).
Additionally, we measure the end-to-end checkpoint loading time (the blocking time of the load API call) for both load-time resharding and standard loading scenarios.
We use the average ETTR as the metric to evaluate the end-to-end system performance.
Following GEMINI~\cite{gemini-sys}, we assume a consistent occurrence of failures within each checkpointing interval (100 steps), calculating the average wasted time to derive the achieved ETTR.
Calculation details can be found in Appendix~\ref{sec:ettr_cal}.

\noindent \textbf{Baselines}.
In experiments with FSDP, we use DCP (commit hash: c7338f4) as the baseline, while for Megatron-LM, MCP~\cite{megatron-dcp} (commit hash: 3fb5c51) serves as the baseline.
We ensure that all hyper-parameters, such as batch size and context length, remain consistent across systems for a fair comparison.
Both baselines support asynchronous checkpointing and load-time checkpoint resharding for the model and optimizer, but not for dataloader states.
Consequently, we compare I/O performance with these baselines only for GPU states and additionally evaluate the performance of \sys{} for the full training states. 
Since neither DCP nor MCP supports HDFS, we implement the 
logic to enable HDFS access for them, 
using the same configurations (e.g., thread count, file chunk size, etc.) as those employed in \sys{}.

\noindent \textbf{Saving performance}.
As shown in Table~\ref{tab:io_compare}, \sys{} significantly reduces checkpoint stalls during runtime, with reductions ranging from 13.06$\times$ to 161.50$\times$. This improvement reduces the average checkpoint stall time from minutes or seconds to sub-second durations, outperforming both DCP and MCP.
In terms of end-to-end saving time, \sys{} delivers an average speedup of 6.05$\times$, primarily driven by our fine-grained performance optimizations.
Notably, the acceleration provided by \sys{} scales with the size of the training workload, increasing from 2.21$\times$ for 2400 GPUs to 8.87$\times$ for 4800 GPUs.
This can be attributed to the workload balancing mechanism, which becomes more effective with larger DP sizes.
Additionally, we observe a particularly significant reduction in blocking time for FSDP workloads. This results from the high communication and synchronization overheads caused by FSDP’s irregular tensor shard processing when using DCP for distributed checkpoint storage. These overheads grow as the training scale increases, making DCP less suitable for large-scale training.
In contrast, \sys{}'s decomposition representation strategy incurs zero communication overhead during metadata generation, leading to improved training efficiency.
We also provide a detailed overhead breakdown in Appendix~\ref{sec:save_breakdown} for further investigation.
For training with Megatron-LM, we report I/O performance for the handling of full training states.

\noindent \textbf{Loading and resharding performance}.
In experiments involving the loading of distributed checkpoints into unchanged parallelism configurations, \sys{} achieves performance improvements ranging from 1.41$\times$ to 8.80$\times$ compared to baseline methods.
Our approach leverages an asynchronous loading pipeline, which overlaps tensor reading and transferring, significantly reducing blocking time before training begins.
When it comes to the resharding efficiency of model and optimizer states, \sys{} delivers an average acceleration of 3.64$\times$.
By eliminating the need for running costly offline resharding jobs, \sys{} enables flexible and efficient checkpoint transformations across various scenarios.
However, we observe that including CPU states, primarily dataloader states, increases the end-to-end load-time resharding time. This is largely due to the time-consuming processing of unique components within the dataloader states, such as the token buffer, which can grow as large as 20GB in text-to-video LFM training.
Additionally, since only a subset of workers hold these dataloader states, they often become stragglers during the checkpoint saving and loading (resharding) procedures.
We leave the design of a more efficient dataloader to our future works.

\noindent \textbf{End-to-end performance}.
\sys{} outperforms both DCP and MCP in terms of the average end-to-end ETTR, achieving improvements ranging from 1.16$\times$ to 1.29$\times$.
Leveraging full-stack optimizations, \sys{} improves the execution efficiency of each checkpoint-related operation, thereby effectively minimizing training downtime.

\begin{table}[!t]
\caption{
Saving optimization microbenchmark.
}
\label{tab:save_bench}
\begin{small}
\resizebox{\linewidth}{!}{
\begin{tabular}{c|c|ccc}
    \toprule
    \textbf{Workload} & \textbf{Parallel Config.} & \textbf{Optimization} & \textbf{Saving Time (s)} \\
    \midrule
     \multirow{4}{*}{tGPT 13B } & \multirow{4}{*}{TP=2, DP=8, PP=2} & No Optim. & 50.26\\
     &  & Async. & 34.68 (\textbf{1.45$\times$})\\
     &  & Async. + WB. & 20.28 (\textbf{2.48$\times$})\\
     &  & Async. + WB. + Cache.  & 19.97 (\textbf{2.52$\times$})\\
     \midrule
     \multirow{4}{*}{tGPT 30B } & \multirow{4}{*}{TP=2, DP=8, PP=4} & No Optim. & 46.34\\
     &  & Async. & 25.56 (\textbf{1.81$\times$})\\
     & & Async. + W.B. & 18.83 (\textbf{2.46$\times$})\\
     &  & Async. + W.B. + Cache. & 18.56 (\textbf{2.50$\times$})\\
    \bottomrule
\end{tabular}}
\end{small}
\end{table}

\begin{table}[!t]
\caption{
Loading optimization microbenchmark.
}
\label{tab:load_bench}
\begin{small}
\resizebox{\linewidth}{!}{
\begin{tabular}{c|c|ccc}
    \toprule
    \textbf{Model} & \textbf{Parallel Config.} & \textbf{Optimization} & \textbf{Loading Time (s)} \\
    \midrule
     \multirow{3}{*}{tGPT 13B } & \multirow{3}{*}{TP=2, DP=8, PP=2} & No Optim. & 63.48\\
     &  & Async. & 48.43 (\textbf{1.31$\times$})\\
     &  & Async. + Overlap. & 41.38 (\textbf{1.53$\times$})\\
     \midrule
     \multirow{3}{*}{tGPT 30B } & \multirow{3}{*}{TP=2, DP=8, PP=4} & No Optim. & 77.02\\
     &  & Async. & 74.54 (\textbf{1.03$\times$})\\
     & & Async. + Overlap. & 48.73 (\textbf{1.58$\times$})\\
    \bottomrule
\end{tabular}}
\end{small}
\end{table}

\label{sec:reshard_optim}
\begin{table}[!t]
\caption{
Resharding optimization microbenchmark.
}
\label{tab:reshard_bench}
\begin{small}
\resizebox{\linewidth}{!}{
\begin{tabular}{c|c|ccc}
    \toprule
    \textbf{Model} & \textbf{Parallel Config.} & \textbf{Optimization} & \textbf{Processing Time (s)} \\
    \midrule
     \multirow{2}{*}{tGPT 13B} & \multirow{2}{*}{ZeRO-2 32 GPUs} & All-gather + D2H. & 4.12\\
     & & Decompose. & 0.21 (\textbf{19.80$\times$})\\
     \midrule
     \multirow{2}{*}{tGPT 30B} &  \multirow{2}{*}{ZeRO-2 64 GPUs} & All-gather + D2H. & 5.84\\
     & & Decompose. & 0.19 (\textbf{30.50$\times$})\\
    \bottomrule
\end{tabular}}
\end{small}
\end{table}

\begin{table*}[!t]
\caption{I/O performance of \sys{} in large-scale LFM training.}
\label{tab:io_product}
\begin{small}
\resizebox{\linewidth}{!}{
\begin{tabular}{c|cc|cccc}
    \toprule
    \textbf{Model and Framework} & \textbf{\#GPUs 
    } & \textbf{Parallelism} & $\mathbf{T_{Block}}$ \textbf{(s)} & $\mathbf{T_{Save}}$ \textbf{(s)} & $\mathbf{T_{Load}}$ \textbf{(s)} \\
    \midrule
     Vision Transformer 7B FSDP & 1488 & ZeRO-2 & 0.34 & 20.13 & 265.73 \\
     Text Transformer 405B Megatron-LM & 8960 & TP=8, DP=70, PP=16 & 0.59 & 51.06 & 129.49 \\
    \bottomrule
\end{tabular}}
\end{small}
\end{table*}

\subsection{Microbenchmarks}
We conduct several micro-benchmarks (Table~\ref{tab:save_bench}, Table~\ref{tab:load_bench} and Table~\ref{tab:reshard_bench}) to highlight the performance improvements.  To evaluate different model sizes,
We modify the tGPT 70B model to create tGPT 13B and tGPT 30B.
For saving and loading microbenchmarks, we used the Megatron-LM training framework, while for resharding, we employed FSDP.

In Table~\ref{tab:save_bench}, we compare the end-to-end saving times with and without \sys{} features, as described in Sec.~\ref{sec:io_optimization}.
The results demonstrate that asynchronous saving pipeline achieves significant reduction in saving time compared to the unoptimized settings, yielding speedups of 1.45$\times$ and 1.81$\times$ for tGPT 13B and tGPT 30B, respectively.
The workload balancing mechanism and plan cache of \sys{} can further boost these improvements.
When applying all I/O optimization techniques to both model and optimizer states, the average saving time for these models is reduced from 48.3 seconds to 19.27 seconds.
In Table~\ref{tab:load_bench}, we present the performance gains achieved by the asynchronous loading pipeline and read-communication overlap techniques.
For tGPT 13B and tGPT 30B, the baseline loading times were 1.53$\times$ and 1.58$\times$ longer than those achieved with \sys{}.

For load-time resharding, we compared the irregular tensor processing time between FSDP and \sys{}'s irregular tensor decomposition strategy. As shown in Table~\ref{tab:reshard_bench}, \sys{} incurs an average blocking time of just 0.20s, which is 25.15$\times$ shorter 
than that of FSDP.
Moreover, our approach consistently achieves microsecond-level blocking times, regardless of the training scale. (see results in Sec.~\ref{sec:exp_main} and Sec.~\ref{sec:exp_product}).

\subsection{Correctness Verification}
\label{sec:exp_correct}

\begin{figure}[!t]
    \centering    
     \begin{subfigure}[b]{0.49\linewidth}
         \centering
         \includegraphics[width=\linewidth]{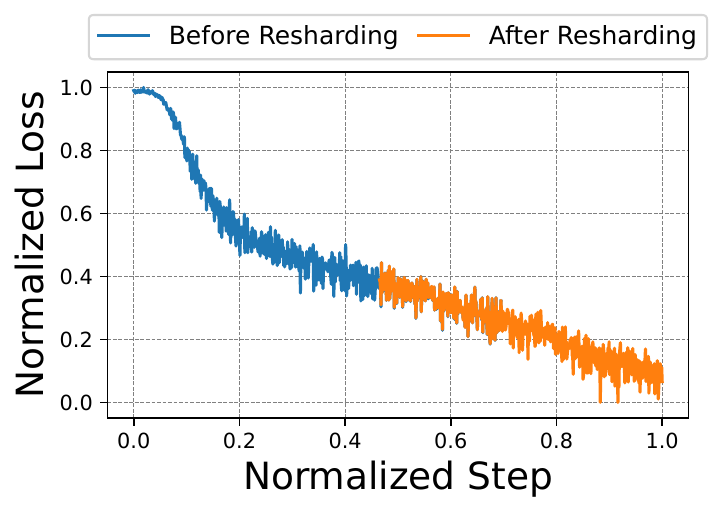}
         \caption{PP resharding from TP=1, DP=4, PP=4 to TP=1, DP=4, PP=8.}
         \label{fig:pp_resharding}
     \end{subfigure}
     \hfill
      \begin{subfigure}[b]{0.49\linewidth}
     \centering
     \includegraphics[width=\linewidth]{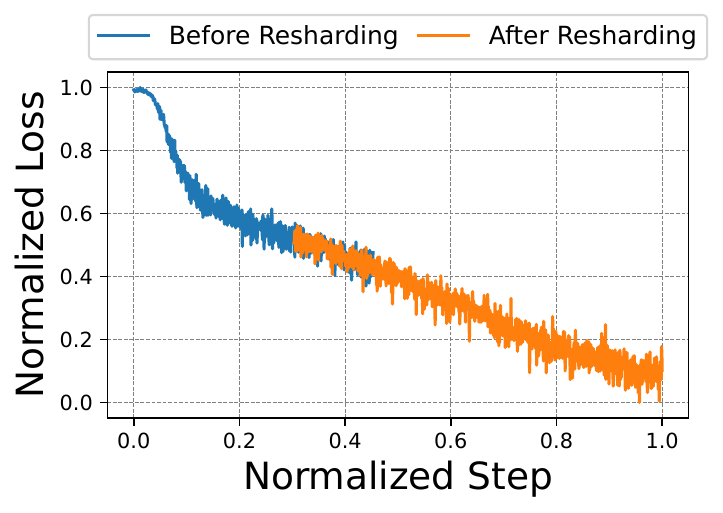}
     \caption{TP resharding from TP=1, DP=4, PP=4 to TP=2, DP=4, PP=4.}
     \label{fig:tp_resharding}
    \end{subfigure}
     \caption{Resharding correctness verification. 
     }
      \label{fig:reshard_correct}
\end{figure}

In Fig.~\ref{fig:reshard_correct}, we depict normalized training loss curves of tGPT 13B before and after resharding with \sys{} in various situations.
For PP and TP resharding, the normalized loss curve after resharding can smoothly match that in the previous phase and continue to display a consistent decline trend.
Unit testing confirms that \sys{} can correctly transform distributed checkpoints across varying parallelism.
Results of DP and hybrid resharding are in Appendix~\ref{sec:exp_full_correct}.

\begin{figure}[!t]
  \centering
  \includegraphics[width=\linewidth]{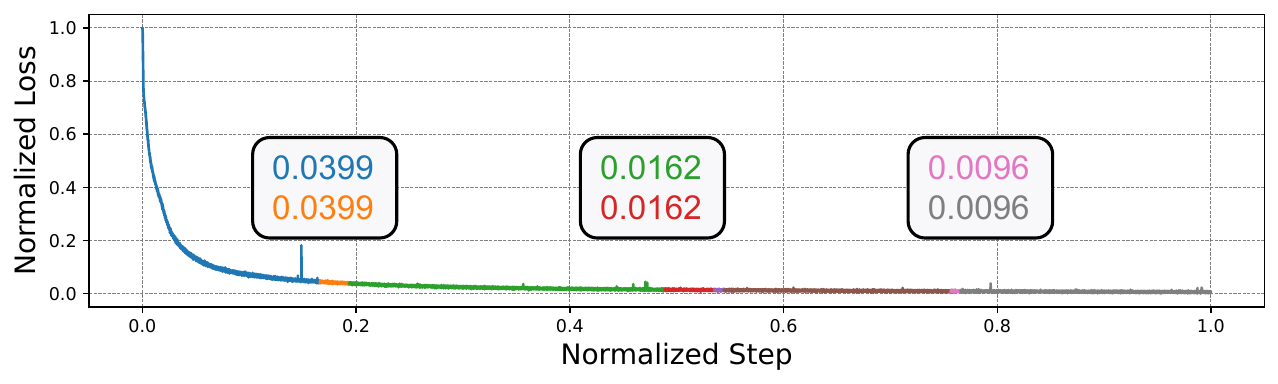}
  \caption{
    Training resuming with \sys{} in real production run on 2080 H800 GPUs 
    for several days.
    Color changes indicate training resuming.
  }
  \vspace{-5mm}
  \label{fig:online_loss}
\end{figure}

We further verify the bit-wise alignment ability of \sys{} when training resumes without changes in parallelism configurations.
We show the normalized loss curve from real production in Fig.~\ref{fig:online_loss}, where a 175B language foundation model is trained.
As highlighted, the normalized loss before and after resuming are exactly the same.
\subsection{Checkpointing in Real LFM Production}
\label{sec:exp_product}
We share our experience deploying \sys{} in real-world LFM production environments, highlighting the issues we encountered and resolved.

\noindent \textbf{Scalabilty}.
As shown in Table~\ref{tab:io_product}, \sys{} was used to train two transformer-based LFMs for image and text generation on H800 GPUs, respectively. 
\sys{} consistently maintained average checkpoint stalls under 600ms, even at the largest scale with 8,960 GPUs.
Additionally, end-to-end checkpointings were completed efficiently, typically within a few seconds.

\noindent \textbf{Dataloader stragglers}. 
By analyzing the performance visualization results from our monitoring tools (Sec.~\ref{sec:visualization}),
we identified prolonged uploading time for dataloader checkpoint files.
To support dataloader state resharding, \sys{} divides the states of a dataloader into multiple parts (e.g., 6 parts for the tGPT experiments in Sec.~\ref{sec:exp_main}) and stores them as individual files.
These files can be independently loaded during resharding.
However, uploading these small files sequentially leads to low bandwidth utilization, causing performance degradation. 
In cases like the 7B vision transformer, uploading time of dataloader states can account for 73.16\% of total saving time.
To fix this issue, we implemented a process pool for concurrent uploads, significantly reducing saving time.

\noindent \textbf{HDFS metadata bottleneck}. 
Through monitoring, we identified several bottlenecks related to HDFS. Despite applying the parallel optimization techniques for single-file uploads described in Sec.~\ref{sec:io_parallel}, the actual HDFS upload performance fell short of our expectations.
A detailed profiling revealed that the metadata concatenation operation in the HDFS NameNode was performed serially, making it a significant performance bottleneck when large volumes of checkpoint files were generated. We mitigated this issue by enabling parallel execution of the concatenation operation.
Another issue that emerged was related to our initial use of the HDFS SDK.
For the convenience of the users, the SDK encapsulates a lot of safeguard logic, leading to unnecessary overheads introduced by checking, creating parent directories, and verifying target files.
To address this, we optimized \sys{} by ensuring directory existence and file uniqueness prior to invoking HDFS APIs, reducing redundant metadata operations and improving performance significantly.
By addressing these issues, we reduced the maximum metadata overhead of HDFS write operations for a single checkpoint file from 3s to  150ms.

\section{Conclusion}
This paper presents the design, implementation, and deployment of \sys{}, a production-grade checkpointing system built for LFM development. 
\sys{} advocates a unified architecture for 
checkpoint representation and workflow, enables efficient load-time  resharding and supports multiple training frameworks and storage backends.
It enables full-stack I/O performance and scalability optimizations and incorporates efficient performance monitoring and analysis tools for LFM development at scale.
Compared to state-of-the-art baselines~\cite{torch-dcp, megatron-dcp}, \sys{} 
reduces checkpoint stalls by up to 161.50$\times$ and shortens end-to-end checkpointing completion time by up to 9.96$\times$. Additionally, the loading (resharding) procedure is optimized to be 3.88$\times$ faster on average.
We believe our system not only provides valuable practical experience to those building checkpointing systems for real-world LFM development but also offers deep insights for future research in the community.
\section{Acknowledgments}\label{sec:ack}
We would like to thank our shepherd Ayush Goel and
the anonymous reviewers for their constructive feedback.
We thank Juncai Liu and Xiang Li for their efforts in facilitating our large-scale experiments.
We thank Shibiao Nong, Shuaishuai Cao, Wei Jia, Jingzhe Tang, Jun Wang, Xi Wang, Sa Wang, and other contributors for refining our system and extending its functionalities.
We thank Liyang Zhao and the HDFS team of ByteDance for their support.
This work was supported in part by a ByteDance Research Collaboration Project, and grants from Hong Kong RGC under the contracts HKU 17204423 and C7004-22G (CRF).

\bibliographystyle{plain}
\bibliography{ref}

\begin{thebibliography}{10}

\bibitem{nnproxy}
{HDFS Federation Solution: NameNodeProxy}.
\newblock \url{https://github.com/bytedance/nnproxy}, 2016.

\bibitem{huggingfaceSafetensors}
{S}afetensors.
\newblock \url{https://huggingface.co/docs/safetensors/index}, 2024.

\bibitem{googleTensorStore}
{T}ensor{S}tore.
\newblock \url{https://google.github.io/tensorstore/}, 2024.

\bibitem{zarrZarr}
{Z}arr.
\newblock \url{https://zarr.dev/}, 2024.

\bibitem{gpt4}
Josh Achiam, Steven Adler, Sandhini Agarwal, Lama Ahmad, Ilge Akkaya, Florencia~Leoni Aleman, Diogo Almeida, Janko Altenschmidt, Sam Altman, Shyamal Anadkat, et~al.
\newblock {GPT-4} technical report.
\newblock {\em arXiv preprint arXiv:2303.08774}, 2023.

\bibitem{fireflyer}
Wei An, Xiao Bi, Guanting Chen, Shanhuang Chen, Chengqi Deng, Honghui Ding, Kai Dong, Qiushi Du, Wenjun Gao, Kang Guan, et~al.
\newblock {Fire-Flyer AI-HPC: A cost-effective software-hardware co-design for deep learning}.
\newblock {\em arXiv preprint arXiv:2408.14158}, 2024.

\bibitem{varuna}
Sanjith Athlur, Nitika Saran, Muthian Sivathanu, Ramachandran Ramjee, and Nipun Kwatra.
\newblock Varuna: scalable, low-cost training of massive deep learning models.
\newblock In {\em Proceedings of the Seventeenth European Conference on Computer Systems}, pages 472--487, 2022.

\bibitem{layernorm}
Jimmy~Lei Ba, Jamie~Ryan Kiros, and Geoffrey~E Hinton.
\newblock Layer normalization.
\newblock {\em arXiv preprint arXiv:1607.06450}, 2016.

\bibitem{gpt3}
Tom Brown, Benjamin Mann, Nick Ryder, Melanie Subbiah, Jared~D Kaplan, Prafulla Dhariwal, Arvind Neelakantan, Pranav Shyam, Girish Sastry, Amanda Askell, et~al.
\newblock Language models are few-shot learners.
\newblock {\em Advances in Neural Information Processing Systems}, 33:1877--1901, 2020.

\bibitem{projectadam}
Trishul Chilimbi, Yutaka Suzue, Johnson Apacible, and Karthik Kalyanaraman.
\newblock Project adam: Building an efficient and scalable deep learning training system.
\newblock In {\em 11th USENIX Symposium on Operating Systems Design and Implementation (OSDI 14)}, pages 571--582, 2014.

\bibitem{deepseekai}
et~al. DeepSeek-AI.
\newblock Deepseek-v2: A strong, economical, and efficient mixture-of-experts language model, 2024.

\bibitem{parcae}
Jiangfei Duan, Ziang Song, Xupeng Miao, Xiaoli Xi, Dahua Lin, Harry Xu, Minjia Zhang, and Zhihao Jia.
\newblock Parcae: Proactive,$\{$Liveput-Optimized$\}$$\{$DNN$\}$ training on preemptible instances.
\newblock In {\em 21st USENIX Symposium on Networked Systems Design and Implementation (NSDI 24)}, pages 1121--1139, 2024.

\bibitem{llama3.1}
Abhimanyu Dubey, Abhinav Jauhri, Abhinav Pandey, Abhishek Kadian, Ahmad Al-Dahle, Aiesha Letman, Akhil Mathur, Alan Schelten, Amy Yang, Angela Fan, et~al.
\newblock The {Llama} 3 herd of models.
\newblock {\em arXiv preprint arXiv:2407.21783}, 2024.

\bibitem{check-n-run}
Assaf Eisenman, Kiran~Kumar Matam, Steven Ingram, Dheevatsa Mudigere, Raghuraman Krishnamoorthi, Krishnakumar Nair, Misha Smelyanskiy, and Murali Annavaram.
\newblock {Check-N-Run}: A checkpointing system for training deep learning recommendation models.
\newblock In {\em 19th USENIX Symposium on Networked Systems Design and Implementation (NSDI 22)}, pages 929--943, 2022.

\bibitem{hdfs}
Apache~Software Foundation.
\newblock {Hadoop Distributed File System}.
\newblock \url{https://hadoop.apache.org/docs/current/hadoop-project-dist/hadoop-hdfs/HdfsDesign.html}, 2024.

\bibitem{serverlessllm}
Yao Fu, Leyang Xue, Yeqi Huang, Andrei-Octavian Brabete, Dmitrii Ustiugov, Yuvraj Patel, and Luo Mai.
\newblock $\{$ServerlessLLM$\}$:$\{$Low-Latency$\}$ serverless inference for large language models.
\newblock In {\em 18th USENIX Symposium on Operating Systems Design and Implementation (OSDI 24)}, pages 135--153, 2024.

\bibitem{grpc}
gRPC Team.
\newblock grpc: A high performance, open source universal rpc framework get started!
\newblock \url{https://grpc.io/}, 2024.

\bibitem{deepseek-r1}
Daya Guo, Dejian Yang, Haowei Zhang, Junxiao Song, Ruoyu Zhang, Runxin Xu, Qihao Zhu, Shirong Ma, Peiyi Wang, Xiao Bi, et~al.
\newblock Deepseek-r1: Incentivizing reasoning capability in llms via reinforcement learning.
\newblock {\em arXiv preprint arXiv:2501.12948}, 2025.

\bibitem{jit-check}
Tanmaey Gupta, Sanjeev Krishnan, Rituraj Kumar, Abhishek Vijeev, Bhargav Gulavani, Nipun Kwatra, Ramachandran Ramjee, and Muthian Sivathanu.
\newblock {Just-In-Time Checkpointing: Low cost error recovery from deep learning training failures}.
\newblock In {\em Proceedings of the Nineteenth European Conference on Computer Systems}, pages 1110--1125, 2024.

\bibitem{unicron}
Tao He, Xue Li, Zhibin Wang, Kun Qian, Jingbo Xu, Wenyuan Yu, and Jingren Zhou.
\newblock Unicron: Economizing self-healing llm training at scale.
\newblock {\em arXiv preprint arXiv:2401.00134}, 2023.

\bibitem{herold2014introduction}
Frank Herold, Sven Breuner, and Jan Heichler.
\newblock {An introduction to BeeGFS}.
\newblock {\em ThinkParQ, Kaiserslautern, Germany, Tech. Rep}, 2014.

\bibitem{characterization}
Qinghao Hu, Zhisheng Ye, Zerui Wang, Guoteng Wang, Meng Zhang, Qiaoling Chen, Peng Sun, Dahua Lin, Xiaolin Wang, Yingwei Luo, et~al.
\newblock Characterization of large language model development in the datacenter.
\newblock In {\em 21st USENIX Symposium on Networked Systems Design and Implementation (NSDI 24)}, pages 709--729, 2024.

\bibitem{gqa}
Drew~A Hudson and Christopher~D Manning.
\newblock {GQA: A new dataset for real-world visual reasoning and compositional question answering}.
\newblock In {\em Proceedings of the IEEE/CVF conference on computer vision and pattern recognition}, pages 6700--6709, 2019.

\bibitem{oobleck}
Insu Jang, Zhenning Yang, Zhen Zhang, Xin Jin, and Mosharaf Chowdhury.
\newblock Oobleck: Resilient distributed training of large models using pipeline templates.
\newblock In {\em Proceedings of the 29th Symposium on Operating Systems Principles}, pages 382--395, 2023.

\bibitem{megascale}
Ziheng Jiang, Haibin Lin, Yinmin Zhong, Qi~Huang, Yangrui Chen, Zhi Zhang, Yanghua Peng, Xiang Li, Cong Xie, Shibiao Nong, et~al.
\newblock {MegaScale}: Scaling large language model training to more than 10,000 {GPUs}.
\newblock In {\em 21st USENIX Symposium on Networked Systems Design and Implementation (NSDI 24)}, pages 745--760, 2024.

\bibitem{adam}
Diederik~P Kingma and Jimmy Ba.
\newblock Adam: A method for stochastic optimization.
\newblock {\em arXiv preprint arXiv:1412.6980}, 2014.

\bibitem{meta-cluster}
Apostolos Kokolis, Michael Kuchnik, John Hoffman, Adithya Kumar, Parth Malani, Faye Ma, Zachary DeVito, Shubho Sengupta, Kalyan Saladi, and Carole-Jean Wu.
\newblock Revisiting reliability in large-scale machine learning research clusters.
\newblock {\em arXiv preprint arXiv:2410.21680}, 2024.

\bibitem{megatron-lm3}
Vijay~Anand Korthikanti, Jared Casper, Sangkug Lym, Lawrence McAfee, Michael Andersch, Mohammad Shoeybi, and Bryan Catanzaro.
\newblock Reducing activation recomputation in large transformer models.
\newblock {\em Proceedings of Machine Learning and Systems}, 5:341--353, 2023.

\bibitem{lyra}
Jiamin Li, Hong Xu, Yibo Zhu, Zherui Liu, Chuanxiong Guo, and Cong Wang.
\newblock Lyra: Elastic scheduling for deep learning clusters.
\newblock In {\em Proceedings of the Eighteenth European Conference on Computer Systems}, pages 835--850, 2023.

\bibitem{easyscale}
Mingzhen Li, Wencong Xiao, Hailong Yang, Biao Sun, Hanyu Zhao, Shiru Ren, Zhongzhi Luan, Xianyan Jia, Yi~Liu, Yong Li, et~al.
\newblock Easyscale: Elastic training with consistent accuracy and improved utilization on gpus.
\newblock In {\em Proceedings of the International Conference for High Performance Computing, Networking, Storage and Analysis}, pages 1--14, 2023.

\bibitem{ddp}
Shen Li, Yanli Zhao, Rohan Varma, Omkar Salpekar, Pieter Noordhuis, Teng Li, Adam Paszke, Jeff Smith, Brian Vaughan, Pritam Damania, et~al.
\newblock {Pytorch distributed: Experiences on accelerating data parallel training}.
\newblock {\em arXiv preprint arXiv:2006.15704}, 2020.

\bibitem{deepspeed-ucp}
Xinyu Lian, Sam~Ade Jacobs, Lev Kurilenko, Masahiro Tanaka, Stas Bekman, Olatunji Ruwase, and Minjia Zhang.
\newblock Universal checkpointing: Efficient and flexible checkpointing for large scale distributed training.
\newblock {\em arXiv preprint arXiv:2406.18820}, 2024.

\bibitem{deepseek-v3}
Aixin Liu, Bei Feng, Bing Xue, Bingxuan Wang, Bochao Wu, Chengda Lu, Chenggang Zhao, Chengqi Deng, Chenyu Zhang, Chong Ruan, et~al.
\newblock Deepseek-v3 technical report.
\newblock {\em arXiv preprint arXiv:2412.19437}, 2024.

\bibitem{datastates-llm}
Avinash Maurya, Robert Underwood, M~Mustafa Rafique, Franck Cappello, and Bogdan Nicolae.
\newblock {DataStates-LLM: Lazy asynchronous checkpointing for large language models}.
\newblock {\em arXiv preprint arXiv:2406.10707}, 2024.

\bibitem{checkfreq}
Jayashree Mohan, Amar Phanishayee, and Vijay Chidambaram.
\newblock {CheckFreq: Frequent, fine-Grained DNN checkpointing}.
\newblock In {\em 19th USENIX Conference on File and Storage Technologies (FAST 21)}, pages 203--216, 2021.

\bibitem{megatron-lm2}
Deepak Narayanan, Mohammad Shoeybi, Jared Casper, Patrick LeGresley, Mostofa Patwary, Vijay Korthikanti, Dmitri Vainbrand, Prethvi Kashinkunti, Julie Bernauer, Bryan Catanzaro, et~al.
\newblock {Efficient large-scale language model training on gpu clusters using Megatron-LM}.
\newblock In {\em Proceedings of the International Conference for High Performance Computing, Networking, Storage and Analysis}, pages 1--15, 2021.

\bibitem{deepfreeze}
Bogdan Nicolae, Jiali Li, Justin~M Wozniak, George Bosilca, Matthieu Dorier, and Franck Cappello.
\newblock Deepfreeze: Towards scalable asynchronous checkpointing of deep learning models.
\newblock In {\em 2020 20th IEEE/ACM International Symposium on Cluster, Cloud and Internet Computing (CCGRID)}, pages 172--181. IEEE, 2020.

\bibitem{rlhf}
Long Ouyang, Jeffrey Wu, Xu~Jiang, Diogo Almeida, Carroll Wainwright, Pamela Mishkin, Chong Zhang, Sandhini Agarwal, Katarina Slama, Alex Ray, et~al.
\newblock Training language models to follow instructions with human feedback.
\newblock {\em Advances in neural information processing systems}, 35:27730--27744, 2022.

\bibitem{tectonic}
Satadru Pan, Theano Stavrinos, Yunqiao Zhang, Atul Sikaria, Pavel Zakharov, Abhinav Sharma, Mike Shuey, Richard Wareing, Monika Gangapuram, Guanglei Cao, et~al.
\newblock Facebook's tectonic filesystem: Efficiency from exascale.
\newblock In {\em 19th USENIX Conference on File and Storage Technologies (FAST 21)}, pages 217--231, 2021.

\bibitem{dit}
William Peebles and Saining Xie.
\newblock Scalable diffusion models with transformers.
\newblock In {\em Proceedings of the IEEE/CVF International Conference on Computer Vision}, pages 4195--4205, 2023.

\bibitem{dpo}
Rafael Rafailov, Archit Sharma, Eric Mitchell, Christopher~D Manning, Stefano Ermon, and Chelsea Finn.
\newblock Direct preference optimization: Your language model is secretly a reward model.
\newblock {\em Advances in Neural Information Processing Systems}, 36, 2024.

\bibitem{zero}
Samyam Rajbhandari, Jeff Rasley, Olatunji Ruwase, and Yuxiong He.
\newblock {ZERO: Memory optimizations toward training trillion parameter models}.
\newblock In {\em SC20: International Conference for High Performance Computing, Networking, Storage and Analysis}, pages 1--16. IEEE, 2020.

\bibitem{gemini1.5}
Machel Reid, Nikolay Savinov, Denis Teplyashin, Dmitry Lepikhin, Timothy Lillicrap, Jean-baptiste Alayrac, Radu Soricut, Angeliki Lazaridou, Orhan Firat, Julian Schrittwieser, et~al.
\newblock Gemini 1.5: Unlocking multimodal understanding across millions of tokens of context.
\newblock {\em arXiv preprint arXiv:2403.05530}, 2024.

\bibitem{diffusion}
Robin Rombach, Andreas Blattmann, Dominik Lorenz, Patrick Esser, and Bj{\"o}rn Ommer.
\newblock High-resolution image synthesis with latent diffusion models.
\newblock In {\em Proceedings of the IEEE/CVF conference on computer vision and pattern recognition}, pages 10684--10695, 2022.

\bibitem{ppo}
John Schulman, Filip Wolski, Prafulla Dhariwal, Alec Radford, and Oleg Klimov.
\newblock Proximal policy optimization algorithms.
\newblock {\em arXiv preprint arXiv:1707.06347}, 2017.

\bibitem{hybridflow}
Guangming Sheng, Chi Zhang, Zilingfeng Ye, Xibin Wu, Wang Zhang, Ru~Zhang, Yanghua Peng, Haibin Lin, and Chuan Wu.
\newblock Hybridflow: A flexible and efficient rlhf framework.
\newblock {\em arXiv preprint arXiv:2409.19256}, 2024.

\bibitem{megatron}
Mohammad Shoeybi, Mostofa Patwary, Raul Puri, Patrick LeGresley, Jared Casper, and Bryan Catanzaro.
\newblock {Megatron-LM: Training multi-billion parameter language models using model parallelism}.
\newblock {\em arXiv preprint arXiv:1909.08053}, 2019.

\bibitem{shvachko2010hdfs}
Konstantin~V. Shvachko.
\newblock {HDFS Scalability: The limits to growth}.
\newblock {\em USENIX Magazine}, 35, 2010.

\bibitem{character-ai}
Character~AI Team.
\newblock Character.ai.
\newblock \url{https://character.ai/}, 2024.

\bibitem{3fs}
DeepSeek Team.
\newblock Fire-flyer file system.
\newblock \url{https://github.com/deepseek-ai/3FS}, 2025.

\bibitem{github-copilot}
Github Team.
\newblock The world’s most widely adopted ai developer tool.
\newblock \url{https://github.com/features/copilot/}, 2024.

\bibitem{megatron-dcp}
Megatron Team.
\newblock Dist checkpointing package.
\newblock \url{https://docs.nvidia.com/megatron-core/developer-guide/latest/api-guide/dist_checkpointing.html}, 2024.

\bibitem{midjourney}
Midjourney Team.
\newblock {Midjourney}.
\newblock \url{https://www.midjourney.com/home}, 2023.

\bibitem{nccl}
NVIDIA~NCCL Team.
\newblock Nvidia collective communications library (nccl).
\newblock \url{https://developer.nvidia.com/nccl}, 2024.

\bibitem{sora}
OpenAI Team.
\newblock Creating video from text.
\newblock \url{https://openai.com/index/sora/}, 2024.

\bibitem{chatgpt}
OpenAI Team.
\newblock {Introducing ChatGPT}.
\newblock \url{https://openai.com/index/chatgpt/}, 2024.

\bibitem{openai-o1}
OpenAI Team.
\newblock {Introducing OpenAI o1}.
\newblock \url{https://openai.com/o1/}, 2024.

\bibitem{torch-dcp}
PyTorch Team.
\newblock {Getting started with Distributed Checkpoint (DCP)}.
\newblock \url{https://pytorch.org/tutorials/recipes/distributed_checkpoint_recipe.html}, 2023.

\bibitem{bark}
Suno Team.
\newblock Bark is a transformer-based text-to-audio model created by suno.
\newblock \url{https://github.com/suno-ai/bark}, 2023.

\bibitem{Zarr}
Zarr Team.
\newblock Zarr: chunked, compressed, n-dimensional arrays.
\newblock \url{https://zarr.dev/}, 2024.

\bibitem{bamboo}
John Thorpe, Pengzhan Zhao, Jonathan Eyolfson, Yifan Qiao, Zhihao Jia, Minjia Zhang, Ravi Netravali, and Guoqing~Harry Xu.
\newblock Bamboo: Making preemptible instances resilient for affordable training of large {DNNs}.
\newblock In {\em 20th USENIX Symposium on Networked Systems Design and Implementation (NSDI 23)}, pages 497--513, Boston, MA, April 2023. USENIX Association.

\bibitem{attention}
A~Vaswani.
\newblock Attention is all you need.
\newblock {\em Advances in Neural Information Processing Systems}, 2017.

\bibitem{vescale}
veScale Team.
\newblock {veScale: A PyTorch Native LLM Training Framework}.
\newblock \url{https://github.com/volcengine/veScale}, 2024.

\bibitem{tenplex}
Marcel Wagenl{\"a}nder, Guo Li, Bo~Zhao, Luo Mai, and Peter Pietzuch.
\newblock Tenplex: Dynamic parallelism for deep learning using parallelizable tensor collections.
\newblock In {\em Proceedings of the ACM SIGOPS 30th Symposium on Operating Systems Principles}, pages 195--210, 2024.

\bibitem{reft}
Yuxin Wang, Shaohuai Shi, Xin He, Zhenheng Tang, Xinglin Pan, Yang Zheng, Xiaoyu Wu, Amelie~Chi Zhou, Bingsheng He, and Xiaowen Chu.
\newblock Reliable and efficient in-memory fault tolerance of large language model pretraining.
\newblock {\em arXiv preprint arXiv:2310.12670}, 2023.

\bibitem{gemini-sys}
Zhuang Wang, Zhen Jia, Shuai Zheng, Zhen Zhang, Xinwei Fu, TS~Eugene Ng, and Yida Wang.
\newblock Gemini: Fast failure recovery in distributed training with in-memory checkpoints.
\newblock In {\em Proceedings of the 29th Symposium on Operating Systems Principles}, pages 364--381, 2023.

\bibitem{wei2021sft}
Jason Wei, Maarten Bosma, Vincent~Y Zhao, Kelvin Guu, Adams~Wei Yu, Brian Lester, Nan Du, Andrew~M Dai, and Quoc~V Le.
\newblock Finetuned language models are zero-shot learners.
\newblock {\em arXiv preprint arXiv:2109.01652}, 2021.

\bibitem{antman}
Wencong Xiao, Shiru Ren, Yong Li, Yang Zhang, Pengyang Hou, Zhi Li, Yihui Feng, Wei Lin, and Yangqing Jia.
\newblock $\{$AntMan$\}$: Dynamic scaling on $\{$GPU$\}$ clusters for deep learning.
\newblock In {\em 14th USENIX Symposium on Operating Systems Design and Implementation (OSDI 20)}, pages 533--548, 2020.

\bibitem{glm}
Aohan Zeng, Xiao Liu, Zhengxiao Du, Zihan Wang, Hanyu Lai, Ming Ding, Zhuoyi Yang, Yifan Xu, Wendi Zheng, Xiao Xia, et~al.
\newblock {GLM-130b: An open bilingual pre-trained model}.
\newblock {\em arXiv preprint arXiv:2210.02414}, 2022.

\bibitem{ppo-framework}
Chi Zhang, Guangming Sheng, Siyao Liu, Jiahao Li, Ziyuan Feng, Zherui Liu, Xin Liu, Xiaoying Jia, Yanghua Peng, Haibin Lin, et~al.
\newblock A framework for training large language models for code generation via proximal policy optimization.

\bibitem{opt}
Susan Zhang, Stephen Roller, Naman Goyal, Mikel Artetxe, Moya Chen, Shuohui Chen, Christopher Dewan, Mona Diab, Xian Li, Xi~Victoria Lin, et~al.
\newblock {OPT: Open pre-trained transformer language models}.
\newblock {\em arXiv preprint arXiv:2205.01068}, 2022.

\bibitem{qsync}
Juntao Zhao, Borui Wan, Yanghua Peng, Haibin Lin, Yibo Zhu, and Chuan Wu.
\newblock {QSync: Quantization-minimized synchronous distributed training across hybrid devices}.
\newblock {\em arXiv preprint arXiv:2407.02327}, 2024.

\bibitem{fsdp}
Yanli Zhao, Andrew Gu, Rohan Varma, Liang Luo, Chien-Chin Huang, Min Xu, Less Wright, Hamid Shojanazeri, Myle Ott, Sam Shleifer, et~al.
\newblock {Pytorch FSDP: Experiences on scaling fully sharded data parallel}.
\newblock {\em arXiv preprint arXiv:2304.11277}, 2023.

\bibitem{swift}
Yuchen Zhong, Guangming Sheng, Juncheng Liu, Jinhui Yuan, and Chuan Wu.
\newblock Swift: Expedited failure recovery for large-scale dnn training.
\newblock In {\em Proceedings of the 28th ACM SIGPLAN Annual Symposium on Principles and Practice of Parallel Programming}, pages 447--449, 2023.

\bibitem{deepseek}
Qihao Zhu, Daya Guo, Zhihong Shao, Dejian Yang, Peiyi Wang, Runxin Xu, Y~Wu, Yukun Li, Huazuo Gao, Shirong Ma, et~al.
\newblock {DeepSeek-Coder-V2: Breaking the Barrier of Closed-Source Models in Code Intelligence}.
\newblock {\em arXiv preprint arXiv:2406.11931}, 2024.

\end{thebibliography}
\appendix
\clearpage
\section{More Background on Maintaining Resharding Scripts}
\label{sec:reshard_scripts}

\begin{figure}[!t]
  \centering
  \includegraphics[width=\linewidth]{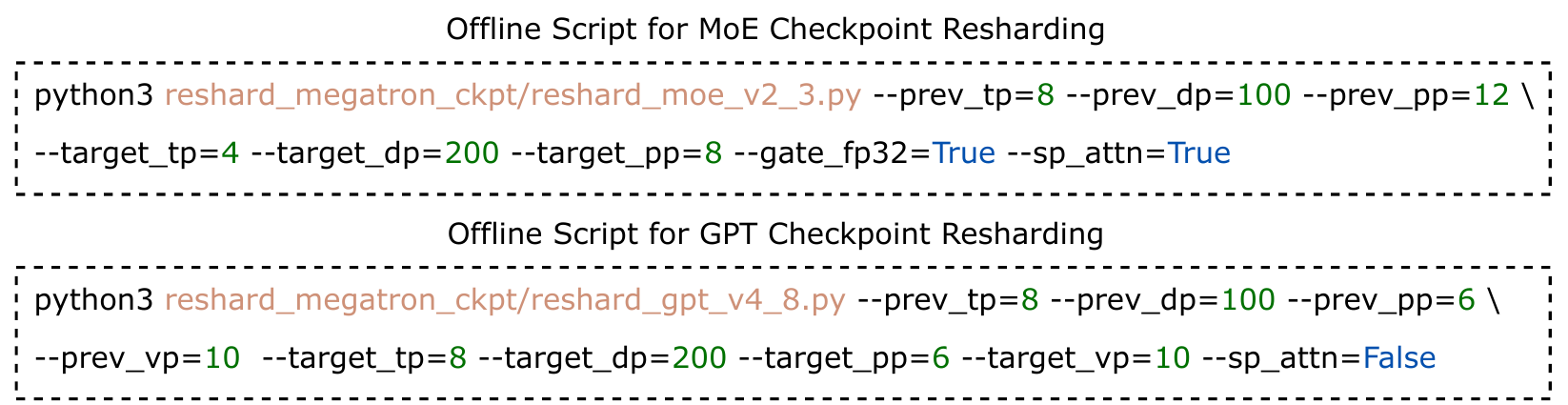}
  \caption{
  Examples of running offline scripts for checkpoint resharding.
  }
  \label{fig:script_example}
\end{figure}

Examples of offline resharding with customized scripts are given in Fig.~\ref{fig:script_example}.
Customizing offline scripts 
is labor-intensive. 
For GPU state resharding, the offline scripts must cover all different components in the LFM and its optimizer and cater to the diverse behavior of each component under different parallelism strategies.
For example, with Tensor Parallelism (TP), 
GEMM operators in attention and MLP blocks are sharded along different dimensions, while other operators like LayerNorm~\cite{layernorm} are replicated across GPUs.
When hybrid 3D parallelism~\cite{megatron-lm2} is adopted, 
TP-sharded tensors of one layer (module) in the distributed optimizer are first flattened and then merged before being sharded according to the designated Data Parallelism (DP) degree.
Offline scripts must implement resharding logic that is tightly coupled with combinations of model (optimizer) components and parallelism strategies. 
Additionally, special algorithmic optimization techniques such as GQA~\cite{gqa} and MLA~\cite{deepseek} change the tensor layout of certain operators 
(e.g., the query-key-value projection GEMM operator in the attention block), necessitating corresponding resharding support.
To handle various cases in our production environment,
our largest script even includes 3193 lines of Python code.
This complexity results in significant engineering efforts for both development and maintenance.

\section{Efficient Integrity Guarantee}
\label{sec:barrier_check}

A complete checkpoint consists of multiple files stored by different workers. The failure of any single worker can corrupt the entire checkpoint.
To prevent such issues, a barrier mechanism is essential to achieve atomic save/load operations among all training workers. 

Checkpointing modules in training frameworks like Megatron-LM~\cite{megatron} rely on the \texttt{barrier} function in \texttt{torch.distributed} to perform integrity checks.
This approach synchronizes training workers to ensure that all checkpoint saving/loading operations are finished.
We observed that when scaling training to involving approximately 10,000 GPUs, this behavior results in stalls of about 20 seconds each time.
To address this inefficiency, we re-implement the \texttt{barrier} function using the aforementioned method (gRPC with tree-based communication topology) 
and conduct the integrity checks asynchronously, 
effectively eliminating the blocking time. We also incorporate upload/download retry mechanisms in \sys{}'s I/O workers
and integrate failure logging, which records the exact stage of failure within the checkpoint saving/loading pipelines in workers who fail to complete checkpointing tasks.

\section{Average ETTR Calculation}\label{sec:ettr_cal}
Failures that happened during training introduce progress loss and the last checkpoint loading overhead.
Assume failures are evenly distributed within one checkpoint interval~\cite{gemini-sys}, the best case is that a failure occurs just after the completion of a checkpoint end-to-end saving procedure whereas the worst case is just before that.
Given the per iteration training time $T_{iter}$, checkpoint interval $N$, end-to-end checkpoint saving time $T_{save}$ and loading (resharding) time $T_{load}$, we derive the average wasted time $T_{wasted}$ as:
\begin{equation}
    T_{wasted} = T_{save} + T_{load} + \frac{N \cdot T_{iter}}{2}
\end{equation}
Therefore, the average ETTR is:
\begin{equation}
    ETTR = 1 - \frac{T_{wasted}}{T_{save} + T_{load} + N\cdot T_{iter}}
\end{equation}

The end-to-end ETTR results presented in Table~\ref{tab:io_compare} are averaged across standard loading and resharding settings.

\section{Checkpointing Overhead Breakdown}
\label{sec:save_breakdown}
We break down the end-to-end checkpoint saving time ($\mathbf{T_{save}}$) for rank 0 by dividing it into several phases and investigating the overhead of each part.
The results are depicted in Table~\ref{tab:save_breakdown}, where $\mathbf{T_{Plan}^{First}}$ denotes the initial planning cost while $\mathbf{T_{Plan}^{Cache}}$ refers to the overhead when using the plan and metadata cache for subsequent checkpointing operations.
Costs of other phases are averaged over various checkpointing steps.
We find that the communication overhead of planning increases as training scales up, leading to significant stalls. 
Thanks to the caching strategy (Sec.~\ref{sec:opt_plan}), it becomes a one-time expense for each (resumed) training session.
In addition, the adoption of the pinned memory CPU pool renders the blocking time of D2H almost negligible, while our asynchronous engine pipeline overlaps the execution time of serialization, shared memory dumping, and HDFS uploading, diminishing the end-to-end time.
Furthermore, the load-balancing mechanism exploits the capabilities for parallel uploading within each DP group and achieves more performance gains at larger training scales (e.g., the uploading speed of model states with 4800 GPUs is 3.03$\times$ faster than that with 2400 GPUs). 

\begin{table*}[!t]
\caption{Detailed overhead breakdown of the checkpoint saving procedure for rank 0.}
\label{tab:save_breakdown}
\begin{small}
\resizebox{\linewidth}{!}{
\begin{tabular}{c|cc|ccccccc}
    \toprule
    \textbf{Model and Framework} & \textbf{Source \#GPUs} & \textbf{State} & $\mathbf{T_{Plan}^{First}}$ \textbf{(s)} & $\mathbf{T_{Plan}^{Cache}}$ \textbf{(s)} & $\mathbf{T_{D2H}}$ \textbf{(s)} & $\mathbf{T_{Serialize}}$ \textbf{(s)} & $\mathbf{T_{Dump}}$ \textbf{(s)} & $\mathbf{T_{Upload}}$ \textbf{(s)}\\
    \midrule
     \multirow{2}{*}{vDiT 4B} & \multirow{2}{*}{32} & Model & 0.05 & 0.00 & 0.15 & 0.52 & 0.50 & 11.04\\
     & & Optimizer & 0.65 & 0.00 & 0.37 & 0.93 & 1.03 & 14.60 \\
     \cline{2-9}
     \multirow{2}{*}{FSDP} & \multirow{2}{*}{128} & Model & 0.35 & 0.00 & 0.06 & 0.29 & 0.12 & 10.57 \\
     & & Optimizer & 0.89 & 0.00 & 0.06 & 0.29 & 0.26 & 9.85 \\
     \midrule
     \multirow{2}{*}{tGPT 70B} & \multirow{2}{*}{2400} & Model & 8.39 & 0.00 & 0.08 & 1.93 & 0.48 & 6.67\\
     & & Optimizer & 5.02 & 0.00 & 0.17 & 2.00 & 2.23 & 2.30 \\
     \cline{2-9}
     \multirow{2}{*}{Megatron-LM} & \multirow{2}{*}{4800}  & Model & 17.09 & 0.00 & 0.08 & 1.95 & 0.47 & 2.20 \\
     &  & Optimizer & 6.46 & 0.00 & 0.03 & 1.69 & 0.35 & 1.67 \\
    \bottomrule
\end{tabular}}
\end{small}
\end{table*}

\section{More Resharding Correctness Experiments}
\label{sec:exp_full_correct}

\begin{figure}[!t]
    \centering    
     \begin{subfigure}[b]{0.49\linewidth}
         \centering
         \includegraphics[width=\linewidth]{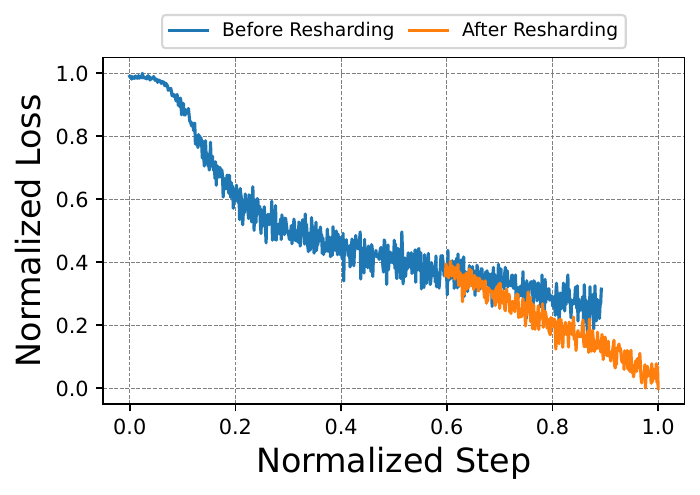}
         \caption{DP resharding from TP=1, DP=4, PP=4 to TP=1, DP=8, PP=4.}
         \label{fig:dp_resharding}
     \end{subfigure}
     \hfill
     \begin{subfigure}[b]{0.49\linewidth}
         \centering
         \includegraphics[width=\linewidth]{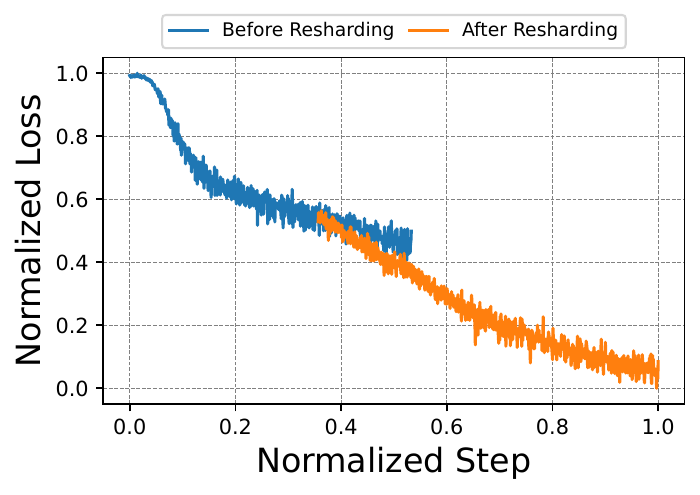}
         \caption{Hybrid resharding from TP=1, DP=4, PP=4 to TP=2, DP=8, PP=2.}
         \label{fig:hp_resharding}
     \end{subfigure}
     \caption{Resharding correctness verification.}
      \label{fig:reshard_full_correct}
\end{figure}

\begin{figure}[!t]
  \centering
  \includegraphics[width=\linewidth]{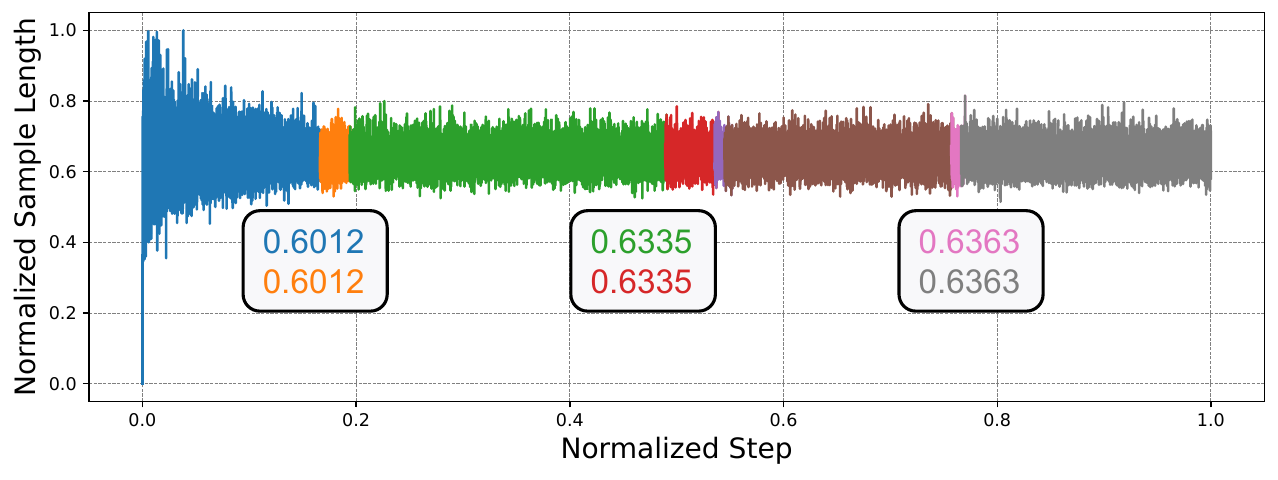}
  \caption{
  The normalized sample length curves of the dataloader with multiple training restarts. 
  }
  \label{fig:loader_resume}
\end{figure}

The normalized loss curves for DP and hybrid resharding correctness are shown in Fig.~\ref{fig:reshard_full_correct}.
In the case of DP and hybrid resharding, since we also increase the global batch size, the loss curve after resharding declines more rapidly.

We further show that \sys{} can achieve bit-wise correct resuming for dataloader states.
Since the RNG state is fixed, correct resumptions should yield the same data sampling trajectory.
Therefore, we use the normalized data sample length curve (Fig.~\ref{fig:loader_resume}) for evaluation.
As highlighted, the normalized data sample length before and after restarts is identical.

\section{Related Work}
\label{sec:related_work}



\noindent \textbf{Checkpointing frameworks.} 
Some industrial initiatives focus on developing checkpointing systems for deep learning. Prior to DCP~\cite{torch-dcp}, PyTorch provided \texttt{torch.save} and \texttt{torch.load} APIs for local checkpoint management, without native resharding support. DCP~\cite{torch-dcp} introduced resharding capabilities for FSDP but lacks support for 
parallelism strategies such as TP and PP. DeepSpeed-UCP~\cite{deepspeed-ucp} provides a unified format for DeepSpeed checkpoints and offers resharding capabilities through offline scripts. Megatron MCP~\cite{megatron-dcp} builds upon the workflow of DCP~\cite{torch-dcp} and extends storage options to formats like Zarr~\cite{Zarr}. 
All these frameworks provide limited support for various parallelism strategies and training frameworks, 
and their I/O performance does not scale well in large-scale training.  

\noindent \textbf{Checkpointing optimizations.} Several 
works \cite{projectadam, deepfreeze} 
 investigated reducing checkpointing costs from different perspectives.
Check-N-Run~\cite{check-n-run}, designed for recommendation models, employs differential checkpointing to store only the modified portions of the model, alongside quantization 
to reduce checkpoint size. 
CheckFreq~\cite{checkfreq} pipelines snapshot and save operations with computation to minimize checkpoint stalls and introduces an online algorithm for tuning checkpoint frequency to further lower costs. 
Gemini~\cite{gemini-sys} advocates for in-memory checkpointing with inter-machine backup for fast recovery, interleaving checkpoint communication with training traffic to enable frequent checkpoints per iteration.
JIT-Checkpointing~\cite{jit-check} adopts just-in-time checkpointing for low-cost error recovery.
In industrial use cases, it is crucial to store checkpoints in separate persistent storage for various tasks such as auto-evaluation~\cite{characterization}, hyper-parameter tuning, and model debugging.
ServerlessLLM~\cite{serverlessllm} focuses on the demand for loading optimizations which is critical in serverless inference scenarios.
It proposes a chunk-based, multi-tier loading pipeline to accelerate checkpoint loading.
Unlike existing solutions, \sys{} champions both optimized I/O performance and flexible load-time checkpoint resharding, supporting general LFM development.

\noindent \textbf{Checkpoint representation.}
PyTorch's \texttt{torch.save/load} functionality relies on \texttt{pickle} for serialization and deserialization.
This format lacks crucial tensor shard metadata, such as global shape information, precluding automatic resharding.
DCP \cite{torch-dcp} addresses this limitation by introducing a disaggregated format that separates metadata from tensor data. This metadata, which includes global shape and offset details, enables automatic resharding in DCP.
Tenplex~\cite{tenplex} introduces Parallelizable Tensor Collection (PTC) to flexibly represent and transform tensor states across diverse parallelism configurations.
To build a PyTorch-native checkpointing system, \sys{} adopts the representation of DCP and incorporates necessary adaptations to handle irregular tensor sharding.
Array-based storage systems like Zarr \cite{zarrZarr} and TensorStore \cite{googleTensorStore} allow tensors to be saved as individual arrays, supporting concurrent reads and writes. MCP \cite{megatron-dcp} supports distributed checkpointing using the Zarr format.
For secure and efficient tensor storage, Safetensors \cite{huggingfaceSafetensors} is a file format for saving tensors safely and loading them efficiently.
To improve compatibility with the Hugging Face open-source ecosystem, \sys{} incorporates functionality to export checkpoints in the Safetensors format.

\noindent \textbf{Storage systems for LFM.}
Large-scale checkpointing requires robust storage systems. The LLaMA 3.1 report \cite{llama3.1} highlights Tectonic \cite{tectonic}, Meta's general-purpose file system, as the backbone for storing checkpoints during pre-training. One of the key challenges for this system is managing high-frequency checkpoint writes. Similarly, FireFlyer \cite{fireflyer}, the AI-HPC system developed by DeepSeek \cite{deepseekai,deepseek-v3,deepseek-r1} for LFM training, utilizes the custom-built 3FS Distributed File System~\cite{3fs}, which is akin to other parallel file systems like BeeGFS~\cite{herold2014introduction}, to manage checkpoint storage. 

\noindent \textbf{Elastic training systems.}
Some orthogonal efforts are devoted to enhancing the elasticity of DL training jobs~\cite{varuna, bamboo, easyscale, parcae, swift, oobleck}.
For instance, Varuna introduces \textit{job morphing} to reconfigure training jobs in both pipeline and data parallelism degrees, without changing the hyper-parameters.
Bamboo~\cite{bamboo} leverages cross-stage redundant computations into the pipeline for elastic training on spot instances.
Oobleck~\cite{oobleck} implements a pipeline instantiation mechanism via predefined templates to tolerate concurrent failures in different pipelines.
These works are limited to specific parallelism strategies (e.g., DP without ZeRO), while \sys{} does not assume any specific parallelism, supporting real-world production.
The flexibility of elastic training systems can be significantly enhanced by adopting our checkpoint representation for training state management.


\end{document}